\documentclass[runningheads]{llncs}
\pdfoutput=1 

\usepackage{times}
\usepackage{epsfig}
\usepackage{graphicx}
\usepackage{amsmath}
\usepackage{amssymb}
\usepackage{amsfonts}
\usepackage{epstopdf}
\usepackage{algorithm}
\usepackage{algpseudocode}
\usepackage{enumitem}
\usepackage{mathtools}
\usepackage{graphics}
\usepackage{subfig}
\usepackage{color,soul}
\usepackage[T1]{fontenc}
\usepackage[font=small,skip=0pt]{caption}
\usepackage{array}
\usepackage{authblk}
\makeatletter
\renewcommand\AB@affilsepx{, \protect\Affilfont}
\makeatother
\newcolumntype{L}[1]{>{\raggedright\arraybackslash\hspace{0pt}}p{#1}}
\newcolumntype{C}[1]{>{\centering\let\newline\\\arraybackslash\hspace{0pt}}p{#1}}
\newcolumntype{R}[1]{>{\raggedleft\let\newline\\\arraybackslash\hspace{0pt}}p{#1}}
\usepackage[width=122mm,left=12mm,paperwidth=146mm,height=193mm,top=12mm,paperheight=217mm]{geometry}
\usepackage[pagebackref=true,breaklinks=true,letterpaper=true,colorlinks,bookmarks=false]{hyperref}
\newcommand*\rot{\rotatebox{0}}

\begin{document}
\pagestyle{headings}
\mainmatter
\def\ECCV16SubNumber{1194}  

\title{Weakly Supervised Learning of Heterogeneous Concepts in Videos} 

\titlerunning{Weakly Supervised Learning of Heterogeneous Concepts in Videos}

\authorrunning{Sohil Shah et al.}

\author{Sohil Shah$^1$, Kuldeep Kulkarni$^2$, Arijit Biswas$^3$, Ankit Gandhi$^3$, Om Deshmukh$^3$ and Larry Davis$^1$}
\institute{$^1$University of Maryland, College Park; $^2$Arizona State University; $^3$Xerox Research Centre India}

\maketitle

\begin{abstract}
Typical textual descriptions that accompany online videos are `weak': i.e., they mention the main concepts in the video but not their corresponding spatio-temporal locations. The concepts in the description are typically heterogeneous (e.g., objects, persons, actions). Certain location constraints on these concepts can also be inferred from the description. The goal of this paper is to present a generalization of the Indian Buffet Process (IBP) that can (a) systematically incorporate heterogeneous concepts in an integrated framework, and (b) enforce location constraints, for efficient classification and localization of the concepts in the videos. Finally, we develop posterior inference for the proposed formulation using mean-field variational approximation. Comparative evaluations on the Casablanca and the A2D datasets show that the proposed approach significantly outperforms other state-of-the-art techniques: 24\% relative improvement for pairwise concept classification in the Casablanca dataset and 9\% relative improvement for localization in the A2D dataset as compared to the most competitive baseline.  
\end{abstract}

\section{Introduction}
\label{sec:introduction}

Watching and sharing videos on social media has become an integral part of everyday life. We are often intrigued by the textual description of the videos and intend to fast-forward to the segments of interest without watching the entire video. However, these textual descriptors usually do not specify the exact segment of the video associated with a particular description. For example, someone describing a movie clip as ``head-on collision between cars while Chris Cooper is driving'' neither provide the time-stamps for the collision or driving events nor the spatial locations of the cars or Chris Cooper. Such descriptions are referred to as `weak labels'. For efficient video navigation and consumption, it is important to automatically determine the spatio-temporal locations of these concepts (such as `collision' or `cars'). However, it is prohibitively expensive to train concept-specific models for all concepts of interest in advance and use them for localization. 
This shortcoming has triggered a great amount of interest in \textit{jointly} learning concept-specific classification models as well as localizing concepts from multiple weakly labeled images \cite{shi2014weakly,leung2011handling,Oquab15} or videos \cite{bojanowski2014weakly,bojanowski2013finding}.


Video descriptions include concepts which may refer to persons, objects, scenes and/or
actions and thus a typical description is a combination of heterogeneous concepts. In the running example, extracted heterogeneous concepts are `car' (object), `head-on collision' (action), `Chris Cooper' (person) and `driving' (action). Learning classifiers for these heterogeneous concepts along with localization is an extremely challenging task because: (a) the classifiers for different kinds of concepts are required to be learned simultaneously, e.g., a face classifier, an object classifier, an action classifier etc., and (b) the learning model must take into account the spatio-temporal location constraints imposed by the descriptions while learning these classifiers. For example, the concepts `head-on collision' and `cars' should spatio-temporally co-occur at least once and there should be at least one car in the video. 

Recently there has been growing interest to jointly learn concept classifiers from weak labels \cite{shi2014weakly,bojanowski2013finding}.  Bojanowski {\it et al} \cite{bojanowski2013finding} proposed a discriminative clustering framework to jointly learn person and action models from movies using weak supervision provided by the movie scripts. Since weak labels are extracted from scripts, each label can be associated with a particular shot in the movie which may last only for a few seconds, i.e., the labels are well localized and that makes the overall learning easier. 
However, in real world videos, one does not have access to such shot-level labels but only to video-level labels. Therefore in our work, we do not assume availability of such well localized labels, and tackle a  more general problem of learning concepts from the weaker video-level labels. The framework in \cite{bojanowski2013finding}, when extended to long videos does not give satisfactory results (see section \ref{sec:experimental_results}). Such techniques, which are based on a linear mapping from features to labels and model background using only a single latent factor, are usually inadequate to capture all the inter-class and intra-class variations.
Shi {\it et al} \cite{shi2014weakly} jointly learn object and attribute classifiers from images using weakly supervised Indian Buffet Process (IBP). Note that IBP \cite{ghahramani2005infinite,griffiths2011indian} 
allows observed features to be explained by a countably infinite number of latent factors. However, the framework in \cite{shi2014weakly} is not designed to handle heterogeneous concepts and location constraints, which leads to a significant degradation in performance (section \ref{sec:res_casa}). \cite{ozdemir2014probabilistic} and \cite{yildirim2012rational} propose IBP based cross-modal categorization/query image retrieval models which learn semantically meaningful abstract features from multimodal (image, speech and text) data. However, these unsupervised approaches do not incorporate any location constraints which naturally arise in the weakly supervised setting with heterogeneous labels.

We propose a novel Bayesian Non-parametric (BNP) approach called WSC-SIIBP (Weakly Supervised, Constrained \& Stacked Integrative IBP) to jointly learn heterogeneous concept classifiers and localize these concepts in videos. BNP models are a class of Bayesian models where the hidden structure that may have generated the observed data is not assumed to be fixed. Instead, a framework is provided that allows the complexity of the model to increase as more data is observed~ \cite{gershman2012tutorial}. Specifically, we propose: 
\vspace{-5mm}
\begin{enumerate}[leftmargin=*,noitemsep]
\item A novel generalization of IBP which for the first time incorporates weakly supervised spatio-temporal location constraints and heterogeneous concepts in an integrated framework.
\item Posterior inference of WSC-SIIBP model using mean-field variational approximation.
\end{enumerate}
\vspace{-2mm}

We assume that the weak video labels come in the form of tuples: in the running example, the extracted heterogeneous concept tuples are (\{car, head-on collision\}, \{Chris Cooper, driving\})\footnote{Extracting the concept tuples from textual descriptions of the videos is an interesting research problem in itself and is beyond the scope of this paper.}. We perform experiments on two video datasets (a) the Casablanca movie dataset \cite{bojanowski2013finding} and (b) the A2D dataset \cite{xu2015can}. We show that the proposed approach WSC-SIIBP outperforms several state-of-the-art methods for heterogeneous concept classification and localization in a weakly supervised setting. 
For example, WSC-SIIBP leads to a relative improvement of 7\%, 5\% and 24\% on person, action and pairwise classification accuracies, respectively, over the most competitive baselines on the Casablanca dataset. Similarly, the relative improvement on localization accuracy is 9\% over the next best approach on the A2D dataset.

\section{Related Work}

\begin{figure}[!tp]
\centering
\includegraphics[width=0.9\textwidth, keepaspectratio]{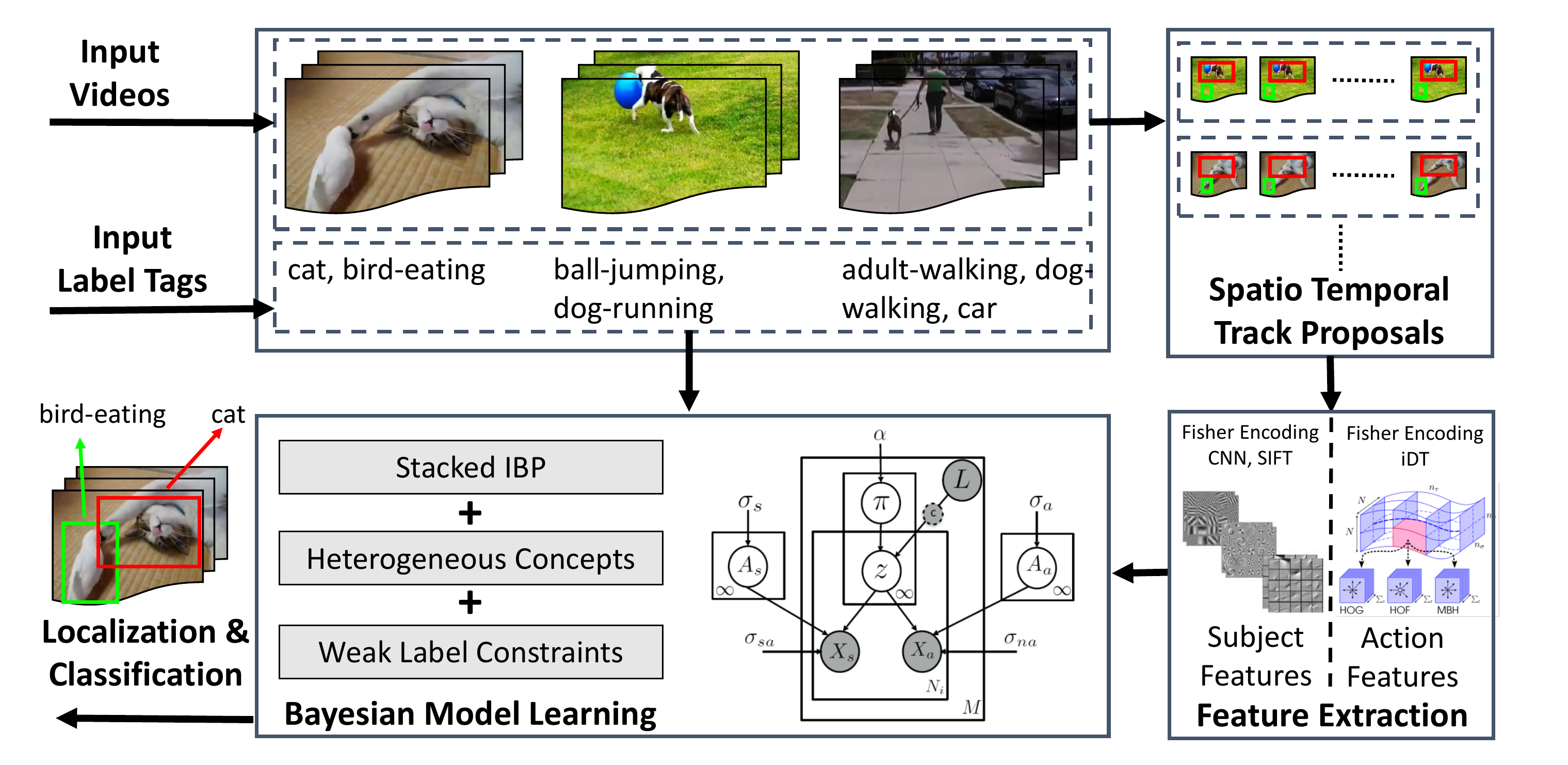}
\caption{Pipeline of WSC-SIIBP. Multiple videos with heterogeneous weak labels are provided as input and localization and classification of the concepts are performed in these videos.}
\label{pipeline}
\vspace{-6mm}
\end{figure}

In this section, we discuss relevant prior work in two broad categories.  

\noindent
{\bf Weakly Supervised Learning:} Localizing concepts and learning classifiers from weakly annotated data is an active research topic. Researchers have learned models for various concepts from weakly labeled videos using Multi-Instance Learning (MIL)~\cite{zhang2005multiple,andrews2002support} for human action recognition\cite{ali2010human}, visual tracking \cite{babenko2009visual} etc. Cour {\it et al} \cite{Cour:eccv08} uses a novel convex formulation to learn face classifiers from movies and TV series using multimodal features which are obtained from finely aligned screenplay, speech and video data. In \cite{bojanowski2015weakly,bojanowski2014weakly}, the authors propose discriminative clustering approaches for aligning videos with temporally ordered text descriptions or predefined tags and in the process also learn action classifiers. In our approach, we consider weak labels which are neither ordered nor aligned to any specific video segment. \cite{prest2012learning} proposes a method for learning object class detectors from real world web videos known to contain only the target class by formulating the problem as a domain adaptation task. \cite{bilen2014object} learns weakly supervised object/action classifiers using a latent-SVM formulation where the objects or actions are localized in training images/videos using latent variables. We note that - both \cite{prest2012learning,bilen2014object} consider only a single weak label per video and unlike our approach, do not jointly learn the heterogeneous concepts. The authors in \cite{tapaswi2015book2movie,zhu2015aligning} use dialogues, scene and character identification to find an optimal mapping between a book and movie shots using shortest path or CRF approach. However, these approaches neither jointly model heterogeneous concepts nor spatio-temporally localized them. Although~\cite{ramanathan2014linking} proposes a discriminative clustering model for coreference resolution in videos, only faces are considered in their experiments.

\noindent
{\bf Heterogeneous concept learning:} There are prior works on automatic image \cite{karpathy2015deep,xu2015show,fang2015captions} and video \cite{venugopalan2015sequence,rohrbach2015long,cho2015describing} caption generation, where models are trained on pairs of image/video and text that contain heterogeneous concept descriptions to predict captions for novel images/videos. While most of these approaches rely on deep learning methods to learn a mapping between an image/video and the corresponding text description, \cite{fang2015captions} uses MIL to learn visual concept detectors (spatial localization in images) for nouns, verbs and adjectives. However, none of these approaches spatio-temporally localize points of interests in videos. Perhaps the available video datasets are not large enough to train such a weakly supervised deep learning model. 

To the best of our knowledge there is no prior work that jointly classifies and localizes heterogeneous concepts in weakly supervised videos.

\section{WSC-SIIBP: Model and Algorithm}

In this section, we describe the details of WSC-SIIBP (see figure \ref{pipeline} for the pipeline). We first introduce notations and motivate our approach in sections \ref{problem_formulation} and \ref{motivation} respectively. This is followed by section \ref{sssec:sibp} where we introduce stacked non-parametric graphical model - IBP and its corresponding posterior computation. In sections \ref{sec:mmibp} and \ref{sec:weak-supervised_constraints}, we formulate an extension of the stacked IBP model which can generalize to heterogeneous concepts as well as incorporate the constraints obtained from weak labels. In section \ref{sec:proposed_inference}, we briefly describe the inference procedure using truncated mean-field variational approximation and summarize our entire algorithm. 
Finally, we discuss how one can classify and localize concepts in new test videos using WSC-SIIBP.

\subsection{Notation}
\label{problem_formulation}

Assume we are given a set of weakly labeled videos denoted by  $\mathbf{\Lambda} = \left\lbrace(i, \Gamma^{(i)})\right\rbrace$, where $i$ indicates a video and $\Gamma^{(i)}$ denotes the heterogeneous weak labels corresponding to the $i$-th video. Although the proposed approach can be used for any number of heterogeneous concepts, for readability, we restrict ourselves to two concepts and call them subjects and actions. We also have a closed set of class labels for these heterogeneous concepts: for subjects $\mathcal{S} = (s_1,\dots,s_{K_s})$ and for actions $\mathcal{A} = (a_1,\dots,a_{K_a})$. Let $K_s = |\mathcal{S}|$, $K_a = |\mathcal{A}|$, $\Gamma^{(i)} = \left\lbrace(s_l,a_l): s_l\in\mathcal{S} \cup \emptyset , a_l \in\mathcal{A} \cup \emptyset , 1\leq l \leq |\Gamma^{(i)}|\right\rbrace$, $\emptyset$ indicate that the corresponding subject or action class label is not present and $M = |\mathbf{\Lambda}|$ represents the number of videos. The video-level annotation simply indicates that the paired concepts $\Gamma^{(i)}$ can occur anywhere in the video and at multiple locations.

Assume that $N_i$ spatio-temporal tracks are extracted from each video \textit{i} where each track \textit{j} is represented as an aggregation of multiple local features, $\mathbf{x}^{(i)}_j$. The spatio-temporal tracks could be face tracks, 3-D object proposals or action proposals (see section \ref{sec:tracks_and_features} for more details). We associate the $j^{th}$ track in video \textit{i} to an infinite binary latent coefficient vector $\mathbf{z}_{j}^{(i)}$ \cite{ghahramani2005infinite,shi2014weakly}. Each video \textit{i} is represented by a bag of spatio-temporal tracks $ \mathbf{X}^{(i)}  = \{\mathbf{x}^{(i)}_j, j = 1,\dots,N_i\}$. Similarly, $ \mathbf{Z}^{(i)}  = \{\mathbf{z}^{(i)}_j, j = 1,\dots,N_i\}$.

\subsection{Motivation}
\label{motivation}

Our objective is to learn (a) a mapping between each of the $N_i$ tracks in video $i$ and the labels in $\Gamma^{(i)}$ and (b) the appearance model for each label identity such that the tracks from new test videos can be classified. To achieve these, it is important for any model to discover the latent factors that can explain similar tracks across a set of videos with a particular label.
In general, the number of latent factors are not known apriori and must be inferred from the data. In Bayesian framework, IBP treats this number as a random variable that can grow with new observations, thus letting the model to effectively explain the unbounded complexity in the data. Specifically, IBP defines a prior distribution over an equivalence class of binary matrices of bounded rows (indicating spatio-temporal tracks) and infinite columns (indicating latent coefficients). To achieve our goals, we build on IBP and introduce WSC-SIIBP model which can effectively learn the latent factors corresponding to each  heterogeneous concept and utilize prior location constraints to reduce the ambiguity in the learning through the knowledge of other latent coefficients.

\vspace{-3mm}
\subsection{Indian Buffet Process (IBP)}
\label{sssec:sibp}

The spatio-temporal tracks in the videos $ \mathbf{\Lambda}$ are obtained from an underlying generative process. Specifically, we consider a stacked IBP model \cite{shi2014weakly} as described below.
\vspace{-1mm}
\noindent
\begin{itemize}[leftmargin=*,noitemsep]
\item For each latent factor $k \in 1 \dots \infty$,
\begin{enumerate}[leftmargin=*,labelindent=.1cm, labelsep=0.1cm,align=left,noitemsep]
\setlength\itemsep{0em}
\item Draw an appearance distribution with mean $\mathbf{a}_{k} \thicksim \mathcal{N}(0,\sigma_A^2\mathbf{I})$
\end{enumerate}
\item For each video $i \in 1\dots M$,
\begin{enumerate}[leftmargin=*,labelindent=.1cm, labelsep=0.1cm,align=left,noitemsep]
\setlength\itemsep{0em}
\item Draw a sequence of i.i.d. random variables, $v^{(i)}_1, v^{(i)}_2 \dots \thicksim$ Beta$(\alpha,1)$
\item Construct the prior on the latent factors, $\pi_k^{(i)} = \prod_{t=1}^k v_t^{(i)}$, $\forall k \in 1\dots \infty,$
\item For $j^{th}$ subject track in $i^{th}$ video, where $j\in 1 \dots N_i$,
\begin{enumerate}[leftmargin=*,itemindent=.5cm,labelwidth=\itemindent,labelsep=0cm,align=left]
\setlength\itemsep{0em}
\item Sample state of each latent factor, $z_{jk}^{(i)} \thicksim$ Bern$(\pi_k^{(i)})$,
\item Sample track appearance, $\mathbf{x}_j^{(i)} \thicksim \mathcal{N}\left(\mathbf{z}_{j}^{(i)}\mathbf{A},\sigma_n^2\mathbf{I}\right)$ 
\end{enumerate}
\end{enumerate}
\vspace{-1mm}
\end{itemize}
where $\alpha$ is the prior controlling the sparsity of latent factors, $\sigma_A^2$ and $\sigma_n^2$ are the prior appearance and noise variance shared across all factors, respectively. Each $\mathbf{a}_{k}$ forms $k^{th}$ row of $\mathbf{A}$ and the value of the latent coefficient $z_{jk}^{(i)}$ indicates whether data $\mathbf{x}_j^{(i)}$ contains the $k^{th}$ latent factor or not. In the above model, we have used stick-breaking construction \cite{doshi2009variational} to generate the $\pi_k^{(i)}$s.
\newline
\noindent
\underline{\bf Posterior}: Now, we describe how the posterior is obtained for the above graphical model. Let $\mathbf{Y} = \left\lbrace\mathbf{\pi}^{(1)}\dots\mathbf{\pi}^{(M)},\mathbf{Z}^{(1)}\dots\mathbf{Z}^{(M)},\mathbf{A}\right\rbrace$ and $\mathbf{\Theta} = \left\lbrace\alpha, \sigma_A^2,\sigma_n^2\right\rbrace$ denote hidden variables and prior parameters, respectively. $\mathbf{X}$ denotes the concatenation of all the spatio-temporal tracks in all $M$ videos, $\left\lbrace\mathbf{X}^{(1)}\dots\mathbf{X}^{(M)}\right\rbrace$. Given prior distribution $\Psi(\mathbf{Y} | \mathbf{\Theta})$ and likelihood function $p(\mathbf{x}^{(i)}_j | \mathbf{Y},\mathbf{\Theta})$, the posterior probability is given by (using Bayes theorem),
\begin{align}
\vspace{-2mm}
&p(\mathbf{Y} | \mathbf{X},\mathbf{\Theta}) = \frac{\Psi(\mathbf{Y} | \mathbf{\Theta}) \prod_{i=1}^M \prod_{j=1}^{N_i}p(\mathbf{X}^{(i)}_j | \mathbf{Y},\mathbf{\Theta})}{p(\mathbf{X} | \mathbf{\Theta})} \label{bayes} \\
&\Psi(\mathbf{Y} | \mathbf{\Theta}) = \prod_{k=1}^\infty \left(\prod_{i=1}^M  p(\pi_k^{(i)} | \alpha) \prod_{j=1}^{N_i} p(z_{jk}^{(i)} | \pi_k^{(i)} )\right) p(\mathbf{a}_{k.} | \sigma_A^2) \nonumber
\end{align}
where $p(\mathbf{X} | \mathbf{\Theta})$ is the marginal likelihood. For simplicity, we denote $p(\mathbf{Y} | \mathbf{X},\mathbf{\Theta})$ as $q(\mathbf{Y})$. Apart from the significance of inferring $\mathbf{Z}^{(i)}$ for identifying track-level labels, inferring prior $\pi_k^{(i)}$ for each video helps to identify video-level labels, while the inference of appearance model $\mathbf{A}$ will be used to classify new test samples (see section \ref{sec:proposed_inference}). Thus, learning in our model requires computing the full posterior distribution over $\mathbf{Y}$.
\newline
\noindent
\underline{\bf Regularized posterior}: We note that it is difficult to infer regularized posterior distributions using Equation \eqref{bayes}. Zellner in \cite{zellner1988optimal} demonstrated that the posterior distribution in \eqref{bayes} can also be obtained as the solution $q(\mathbf{Y})$ of the following optimization problem,
\begin{equation}
{\small
\begin{aligned}
\label{eq3.5}
\min_{q(\mathbf{Y})} \quad &\text{KL}\left(q(\mathbf{Y}) || \Psi(\mathbf{Y}|\mathbf{\Theta})\right)
 - \sum_{i=1}^M \sum_{j=1}^{N_i} \int \log p(\mathbf{x}^{(i)}_j | \mathbf{Y}, \mathbf{\Theta}) q(\mathbf{Y}) d\mathbf{Y} \\
 \vspace{-4mm}
s.t. \quad &q(\mathbf{Y}) \in P_{prob}
\end{aligned}
}
\end{equation}
where $\text{KL}(.)$ denotes the Kullback-Liebler divergence and $P_{prob}$ is the probability simplex. As we will see later, this procedure enables us to learn posterior distribution using constrained optimization framework.

\subsection{Integrative IBP}
\label{sec:mmibp}

Our objective is to model heterogeneous concepts (such as subjects and actions) using a graphical model. However, the IBP model described above can not handle multiple concepts because it is highly unlikely that the subject and the action features can be explained by the same statistical model. Hence, we propose an extension of stacked IBP for heterogeneous concepts, where different concept types are modeled using different appearance models.

Let the subject and action types corresponding to the spatio-temporal track \textit{j} in video \textit{i} be denoted by $\mathbf{x^s}^{(i)}_j$ and $\mathbf{x^a}^{(i)}_j$, respectively, with each having different dimensions $D^e$ ($e \in \{s,a\}$)\footnote{We often use $e$ as a replacement of $s$ and $a$ throughout the paper.}. Unlike the IBP model, $\mathbf{X^s}^{(i)}_j$ and $\mathbf{X^a}^{(i)}_j$ are now represented using two different gaussian noise models $\mathcal{N}(\mathbf{z}^{(i)}_{j}\mathbf{A}^s, \sigma_{ns}^2\mathbf{I})$ and $\mathcal{N}(\mathbf{z}^{(i)}_{j}\mathbf{A}^a, \sigma_{na}^2\mathbf{I})$ respectively where $\sigma_{ne}^2$ denotes prior noise variance and $\mathbf{A}^e$ are $K \times D^e$ matrices (K $\rightarrow \infty$). The mean of the subject and action appearance models for each latent factor are also sampled independently from gaussian distributions of different variances $\sigma_{Ae}^2$. The new posterior probability is given by,
\begin{equation}
\begin{aligned}
\label{eq3.3a}
\tilde{q}(\mathbf{Y}) &= \frac{\Psi(\mathbf{Y} | \mathbf{\Theta}) \prod_{i=1}^M \prod_{j=1}^{N_i}\prod_{e\in\{s,a\}}p(\mathbf{x^e}^{(i)}_j | \mathbf{Z},\mathbf{A}^e, \mathbf{\Theta})}{p(\mathbf{X} | \mathbf{\Theta})} \\
{\Psi}(\mathbf{Y} | \mathbf{\Theta}) &= \prod_{k=1}^\infty \left(\prod_{i=1}^M  p(\pi_k^{(i)} | \alpha) \prod_{j=1}^{N_i} p(z_{jk}^{(i)} | \pi_k^{(i)} )\right) \prod_{\mathclap{e\in\{s,a\}}} p(\mathbf{a}^e_{k} | \sigma_{Ae}^2\mathbf{I})
\end{aligned}
\end{equation}

\subsection{Integrative IBP with Constraints}
\label{sec:weak-supervised_constraints}

Although the graphical model described above is capable of handling heterogeneous features, the location constraints inferred from the weak labels still need to be incorporated into the graphical model. As motivated in section \ref{sec:introduction}, 
the concepts `head-on collision' and `cars' should spatio-temporally co-occur at least once and there should be at least one car in the full video. Imposing these location constraints in the inference algorithm can lead to more accurate parameter estimation of the graphical model and faster convergence of the inference procedure. These constraints can be generalized as follows,
\vspace{-2mm}
\begin{enumerate}[leftmargin=*,labelindent=.1cm, labelsep=0.1cm,align=left,noitemsep]
\setlength\itemsep{0em}
\item Every label tuple in $\Gamma^{(i)}$, is associated with at least one spatio-temporal track (i.e., the event occurs in the video).
\item Spatio-temporal tracks should be assigned a label only from the list of weak labels assigned to the video. Concepts present in the video but not in the label will be subsumed in the background models.
\end{enumerate}
\vspace{-2mm}
Ideally in the case of noiseless labels, these constraints should be strictly followed. However, we assume that real-world labels could be noisy and noise is independent of the videos. Hence, we allow constraints to be violated but penalize the violations using additional slack variables.

We associate the first $K_s$ and the following $K_a$ latent factors (the rows of $\mathbf{A}$) to the subject and action classes in $\mathcal{S}$ and $\mathcal{A}$ respectively. The inferred values of their corresponding latent coefficients in $\mathbf{z}^{(i)}_{j}$ are used to determine the presence/absence of the associated concept in a particular spatio-temporal track. The remaining unbounded number of latent factors are used to explain away the background tracks from unknown action and subject classes in a video. With these assignments, we enforce the following constraints on latent factors which are sufficient to satisfy the conditions mentioned earlier.

To satisfy 1, we introduce the following constraints, $\forall i \in 1\dots M,$ and $\forall j \in 1\dots N_i$,
\vspace{-2mm}
\begin{align}
\vspace{-4mm}
& \sum_{j=1}^{N_i} z_{js}^{(i)} z_{ja}^{(i)} \geq 1 - \xi_{(s,a)}^{(i)}, \quad \forall (s,a)\in \Gamma^{(i)} \label{eq1} \\[-0.5em]
\vspace{-2mm}
& \sum_{j=1}^{N_i} z_{js}^{(i)} \geq 1 - \xi_{(s,\emptyset)}^{(i)}, \quad \forall (s,\emptyset)\in \Gamma^{(i)} \label{eq2}\\[-0.5em]
\vspace{-2mm}
& \sum_{j=1}^{N_i}  z_{ja}^{(i)} \geq 1 - \xi_{(\emptyset,a)}^{(i)}, \quad \forall (\emptyset,a)\in \Gamma^{(i)} \label{eq3}
\end{align}
where $\xi$ is the slack variable, $z_{js}$ and $z_{ja}$ are the latent factor coefficients corresponding to subject class $s$ and action class $a$ respectively. 

To satisfy 2, we use the following constraints, $\forall i \in 1\dots M$ and $\forall j \in 1\dots N_i$,
\vspace{-1mm} 
\begin{align}
 z_{js}^{(i)} &= 0, \text{if } \nexists (s,\emptyset)\in\Gamma^{(i)} \text{ and } \nexists (s,a) \in \Gamma^{(i)}, \forall a \in \mathcal{A}\label{eq4}\\
 z_{ja}^{(i)} &= 0,  \text{if } \nexists (\emptyset,a)\in\Gamma^{(i)} \text{ and } \nexists  (s,a) \in \Gamma^{(i)},  \forall s \in \mathcal{S} \label{eq5}
\vspace{-2mm} 
\end{align}
\vspace{-6mm} 

\begin{figure}[!tp]
\begin{minipage}[c]{0.57\textwidth}
\includegraphics[height=0.30\textheight]{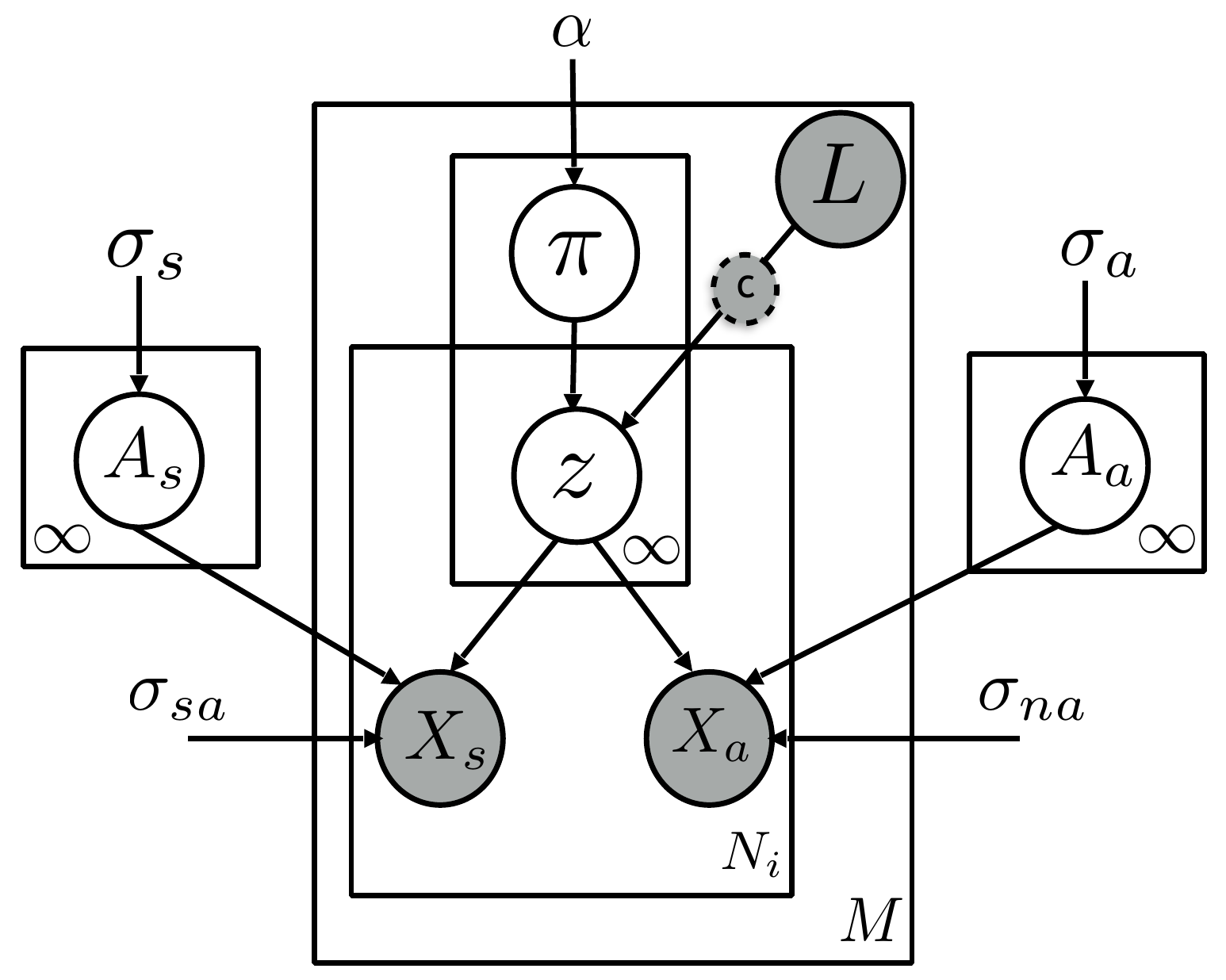}
\end{minipage}\hfill
\begin{minipage}[c]{0.4\textwidth}
\caption{WSC-SIIBP: Graphical Model using two heterogeneous concepts, subjects and actions. Each video (described by video-level labels $L$) is independently modeled using latent factor prior $\pi$ and contains $N_i$ tracks. Each track is represented using subject and action features $X_s$ and $X_a$ respectively, which are modeled using Gaussian appearance models $A_s$ and $A_a$. $z$ are the binary latent variables indicating the presence or absence of the latent factors in each track. $c$ denotes the set of location constraints extracted from the video labels.}
\label{fig2}
\end{minipage}
\end{figure}

The constraints defined in \eqref{eq1}-\eqref{eq5} have been used in the context of discriminative clustering ~\cite{ramanathan2014linking,bojanowski2013finding}. However, our model is the first to use these constraints in a Bayesian setup. In their simplest form, they can be enforced using the  point estimate of $z$ e.g., MAP estimation. However, $\mathbf{Z}^{(i)}$ is defined over the entire probability space. To enforce the above constraints in a Bayesian framework, we need to account for the uncertainty in $\mathbf{Z}^{(i)}$. Following \cite{zhu2014bayesian,ganchev2010posterior}, we define effective constraints as an expectation of the original constraints in \eqref{eq1}-\eqref{eq5}, where the expectation is computed w.r.t.\ the posterior distribution in \eqref{eq3.3a} (see supplementary material for the expectation constraints). The proposed graphical model, incorporating heterogeneous concepts as well as the location constraints provided by the weak labels, is shown in figure \ref{fig2}. 

We restrict the search space for posterior distribution in Equation \eqref{eq3.3a} by using the expectation constraints. In order to obtain the regularized posterior distribution of the proposed model, we solve
the following optimization problem under these expectation constraints,
\vspace{-3mm}
\begin{align}
\min_{\tilde{q}(\mathbf{Y}), \mathbf{\xi}^{(i)}} &\quad \text{KL}\left(\tilde{q}(\mathbf{Y}) || \tilde{\Psi}(\mathbf{Y}|\mathbf{\Theta})\right) &-& \sum_{i=1}^M \sum_{j=1}^{N_i} \int \left(\sum_{e\in\{s,a\}}\log p\left(\mathbf{X^e}^{(i)}_j | \mathbf{Y}, \mathbf{\Theta}\right) \right) \tilde{q}(\mathbf{Y}) d\mathbf{Y} \nonumber \\
& & + & C \sum_{i=1}^M \sum_{J \in \Gamma^{(i)}} \xi^{(i)}_J \nonumber \\
s.t. \quad & \tilde{q}(\mathbf{Y}) \in P_{prob} && \label{eq11}
\vspace{-3mm}
\end{align}

\subsection{Learning and Inference}
\label{sec:proposed_inference}

Note that the variational inference for true posterior $\tilde{q}(\mathbf{Y})$ (in Equation (\ref{eq3.3a})) is intractable over the general space of probability functions. To make our problem easier to solve, we establish \textit{truncated} mean-field variational approximation~\cite{doshi2009variational} to the desired posterior $\tilde{q}(\mathbf{Y})$, such that the search space $P_{prob}$ is constrained by the following tractable parametrised family of distributions,
\begin{equation}
{\small
\begin{aligned}
\label{eq12}
\tilde{w}(\mathbf{Y}) =& \prod_{i=1}^M \left(\prod_{k=1}^{K_{max}} p(v_k^{(i)} | \tau_{k1}^{(i)}, \tau_{k2}^{(i)}) \prod_{j=1}^{N_i} p(z_{jk}^{(i)} | \nu_{jk}^{(i)}) \right)
\prod_{k=1}^{K_{max}} \prod_{\mathclap{\,\,\,\,\,\,e\in\{s,a\}}} p(\mathbf{a}^e_{k}|\mathbf{\Phi}_k^e, \sigma_{ke}^2\mathbf{I}) 
\end{aligned}
}
\end{equation}

where $p(v_k^{(i)} | \tau_{k1}^{(i)}, \tau_{k2}^{(i)}) = \text{Beta}(v_k^{(i)}; \tau_{k1}^{(i)}, \tau_{k2}^{(i)})$, $p(z_{jk}^{(i)} | \nu_{jk}^{(i)}) = \text{Bern}(z_{jk}^{(i)}; \nu_{jk}^{(i)})$ and $p(\mathbf{a}^e_{k}|\mathbf{\Phi}_k^e, \sigma_{ke}^2\mathbf{I}) = \mathcal{N}(\mathbf{a}^e_{k}; \mathbf{\Phi}_k^e, \sigma_{ke}^2\mathbf{I})$. In Equation (\ref{eq12}), we note that all the latent variables are modeled independently of all other variables, hence simplifying the inference procedure. The truncated stick breaking process of $\pi_k^{(i)}$'s is bounded at $K_{max}$, wherein $\pi_k = 0$ for $k > K_{max} \gg K_s + K_a + K_{bg}$. $K_{bg}$ indicates the number of latent factors chosen to explain background tracks. 

The optimization problem in Equation \eqref{eq11} is solved using the posterior distribution from Equation \eqref{eq12}. We obtain the parameters (see supplementary material for details) $\sigma_{ke}^2, \mathbf{\Phi^e}_k$, $\tau_{ke}^{(i)}$ and $\nu_{jk}^{(i)}$ for the optimal posterior distribution $\tilde{q}(\mathbf{Y})$ using iterative update rules as summarized in Algorithm \ref{inference}. 
The mean of binary latent coefficients $z_{jk}$, denoted by $\nu_{jk}$, has an update rule which will lead to several interesting observations.
%
\begin{align}
\nu_{jk}^{(i)} &= \frac{L_k^{(i)}}{1+e^{-\zeta_{jk}^{(i)}}}  \label{eq24}
\end{align}
\vspace{-4mm}
\begin{equation}
\small{
\begin{aligned}
 \label{eq25}
\zeta_{jk}^{(i)} &= \sum_{j=1}^k \left(\Psi(\tau_{j1}^{(i)}) - \Psi(\tau_{j1}^{(i)} + \tau_{j2}^{(i)})\right) - \mathcal{L}_k - \sum_{e\in\{s,a\}}\frac{1}{2\sigma_{ne}^2}\left(D^e\sigma_{ke}^2 + \mathbf{\Phi^e}_k\mathbf{\Phi^e}_k^T\right)  \\ 
&+ \sum_{e\in\{s,a\}} \frac{1}{\sigma_{ne}^2}  \mathbf{\Phi^e}_k\left(\mathbf{x}_j^{(i)} - \sum_{l\neq k} \nu_{jl}^{(i)}\mathbf{\Phi^e}_l\right)^T 
+ C\underbrace{\sum_{\substack{J\in \Gamma^{(i)} \\ J=(k,a)}} \mathbb{I}_{\left\lbrace\sum_{l=1}^{N_i}\nu_{lk}^{(i)}\nu_{la}^{(i)} < 1\right\rbrace} \nu_{ja}^{(i)}}_\text{(i)} \\
 &+ C\overbrace{\sum_{\substack{J\in \Gamma^{(i)} \\ J=(s,k)}} \mathbb{I}_{\left\lbrace\sum_{l=1}^{N_i}\nu_{ls}^{(i)}\nu_{lk}^{(i)} < 1\right\rbrace} \nu_{js}^{(i)}}^\text{(ii)} + C\overbrace{\mathbb{I}_{\left\lbrace\sum_{l=1}^{N_i}\nu_{lk}^{(i)} < 1, k \leq K_a + K_s\right\rbrace}}^\text{(iii)}
 \end{aligned}
 }
\end{equation}

\noindent
where $\Psi(.)$ is the digamma function, $\mathbb{I}$ is an indicator function, $L_k^{(i)}$  is an indicator variable and $\mathcal{L}_k $ is a lower bound for $\mathbb{E}_{\tilde{w}}[\log(1 - \prod_{j=1}^k v^{(i)})]$. The $L_k^{(i)}$ indicates whether a concept (action/subject) $k$ is part of $i^{th}$ video label set $\Gamma^{(i)}$ or not. If $L_k^{(i)} = 0$, all the corresponding binary latent coefficients $z_{jk}^{(i)}, j = \{1,\dots,N_i\}$, are forced to 0, which is equivalent to enforcing the constraints in Equation \eqref{eq4} and \eqref{eq5}. Note that the value of $\nu_{jk}^{(i)}$ increases with $\zeta_{jk}^{(i)}$. The terms (i)-(iii) in the update rule for $\zeta_{jk}^{(i)}$ (Equation (\ref{eq25})), which are obtained due to the location constraints in Equation \eqref{eq1}-\eqref{eq3}, act as the coupling terms between $\nu_{je}^{(i)}$'s. For example, for any action concept, term (ii) suggests that if the location constraints are not satisfied, better localization of all the coupled subject concepts (high value of $\nu_{js}^{(i)}$) will drive up the value of $\zeta_{ja}^{(i)}$. This implies that the strong localization of one concept can lead to better localization of other concepts.

The hyperparameter $\sigma_{ne}^2$ and $\sigma_{Ae}^2$ can be set apriori or estimated from data. Similar to the maximization step of EM algorithm, their empirical estimation can easily be obtained by maximizing the expected log-likelihood (see supplementary material). 

\begin{algorithm}[!htp]
\caption{Learning Algorithm of WSC-SIIBP}
\label{inference}
\begin{algorithmic}[1]
\State \textbf{Input:} \parbox[t]{\dimexpr\linewidth-\algorithmicindent} {data $\mathbf{\Lambda} = \lbrace (i, \mathbf{\Gamma}^{(i)})\rbrace_{i\in 1\dots M}$, constant $\alpha, K_{max}$, $C$\strut}
\State \textbf{Output:} \parbox[t]{\dimexpr\linewidth-\algorithmicindent} {distribution $p(\mathbf{v}), p(\mathbf{Z}), p(\mathbf{A}^s), p(\mathbf{A}^a)$ and hyper-parameters $\sigma_{ns}^2, \sigma_{na}^2, \sigma_{As}^2$ and $\sigma_{Aa}^2$\strut}
\State \textbf{Initialize:} \parbox[t]{\dimexpr\linewidth-\algorithmicindent} {$\tau_{k1}^{(i)} = \alpha, \tau_{k2}^{(i)} = 1, \nu_{jk}^{(i)} = 0.5, \mathbf{\Phi}_k^s = \mathbf{\Phi}_k^a = 0, \sigma_{ks}^2 = \sigma_{ka}^2 = \sigma_{ns}^2 = \sigma_{na}^2 = \sigma_{As}^2  = \sigma_{Aa}^2 = 1$\strut}
\Repeat
\Repeat
\State \parbox[t]{\dimexpr\linewidth-\algorithmicindent} {update $\sigma_{ke}^2$ and $\mathbf{\Phi^e}_{k.}$ 
, $\forall 1\leq k \leq K_{max}$, $e \in \{s,a\}$\strut;}
\State \parbox[t]{\dimexpr\linewidth-\algorithmicindent} {update $\tau_{k1}^{(i)}$ and $\tau_{k2}^{(i)}$ 
, $\forall 1\leq k \leq K_{max}$ and $i\in$ 1 to M;\strut}
\State \parbox[t]{\dimexpr\linewidth-\algorithmicindent} {update $\nu_{jk}^{(i)}$ using Equation \eqref{eq24} and \eqref{eq25}, $\forall 1\leq k \leq K_{max}, 1\leq j \leq N_i$ and $i\in$ 1 to M;\strut}
\Until{\parbox[t]{\dimexpr\linewidth-\algorithmicindent} { T iterations or $\frac{\parallel L(t-1) - L(t)\parallel}{L(t)} \leq 1e^{-3}$\strut}}
\State \parbox[t]{\dimexpr\linewidth-\algorithmicindent} {update the hyperparameters $\sigma_{As}^2$, $\sigma_{Aa}^2$, $\sigma_{ns}^2$, $\sigma_{na}^2$ \strut}
\Until{\parbox[t]{\dimexpr\linewidth-\algorithmicindent} { T' iterations or $\frac{\parallel L(t'-1) - L(t')\parallel}{L(t')} \leq 1e^{-4}$\strut}}
\end{algorithmic}
\end{algorithm}

Given the input features $\mathbf{X_s}$ and $\mathbf{X_a}$, the inferred latent coefficients $\nu_{je}^{(i)}$ estimate presence/absence of associated classes in a video. One can classify each spatio-temporal track by estimating the track-level labels using $L^*_j = \arg\max_{k} \nu_{jk}$. Here the maximization is over the latent coefficients corresponding to either the subject or action concepts depending upon the label which we are interested in extracting. For the concept localization task in a video with label pair $(s,a)$, the best track in the video is selected using $j^* = \arg\max_{j} \nu_{js}\times\nu_{ja}$.

\noindent
{\bf Test Inference}: Although the above formulation is proposed for concept classification and localization in a given set of videos (transductive setting), the same algorithm can also be applied to unseen test videos. The latent coefficients for the tracks of test videos can be learned alongside the training data except that the parameters $\sigma_{ke}^2$, $\mathbf{\Phi^e}_{k.}$, $\sigma_{Ae}^2$ and $\sigma_{ne}^2$ are updated only using training data. In the case of free annotation, i.e., absence of labels for test video $i$, we run the proposed approach by setting $L^{(i)}_k = 1$ in eq \eqref{eq24}, indicating that the tracks in a video $i$ can belong to any of the classes in $\mathcal{S}$ or $\mathcal{A}$ (i.e., no constraints as defined by \eqref{eq1}-\eqref{eq5} are enforced). 

\section{Experimental Results}
\label{sec:experimental_results}

In this section, we present an evaluation of WSC-SIIBP on two real-world databases: Casablanca movie and A2D dataset, which represent typical `in-the-wild' videos with weak labels on heterogeneous concepts.

\subsection{Datasets}
\label{sec:tracks_and_features}

\noindent
\underline{\bf Casablanca dataset}: This dataset, introduced in \cite{bojanowski2013finding}, has 19 persons (movie actors) and three action classes (sitdown, walking, background). The heterogeneous concepts used in this dataset are persons and actions. The Casablanca movie is divided into shorter segments of duration either 60 or 120 seconds. We manually annotate all the tracks in each video segment which may contain multiple persons and actions.
Given a video segment and the corresponding video-level labels (extracted from all ground truth track labels), our algorithm maps each of these labels to one or more tracks in that segment, i.e., converts the weak labels to strong labels. Our main objective of evaluation on this dataset is to compare the performance of various algorithms in classifying tracks from videos of varying length.

For our setting, we consider face and action as the two heterogeneous concepts and thus it is required to extract the face and the corresponding action track features. We extract 1094 facial tracks from the full 102 minute Casablanca video. The face tracks are extracted by running the multi-view face detector from  \cite{ZhuR_CVPR_2012} in every frame and associating detections across frames using point tracks~\cite{everingham2006hello}. We follow \cite{parkhi2014compact} to generate the face track feature representations: Dense rootSIFT features are extracted for each face in the track followed by PCA and video-level Fisher vector encoding. The action tracks corresponding to 1094 facial tracks are obtained by extrapolating the face bounding-boxes using linear transformation~\cite{bojanowski2013finding}. For action features, we compute Fisher vector encoding on dense trajectories \cite{wang2013action} extracted from each action track. 

On an average, each 60 sec. segment contains 11 face-action tracks and 4 face-action annotations while each 120 sec. video contains 21 tracks and 6 annotations. Note that, our experimental setup is  more difficult compared to the experimental setting considered in \cite{bojanowski2013finding}. In \cite{bojanowski2013finding}, the Casablanca movie is divided into numerous bags based on the movie script, where on average each segment is of duration 31 sec. containing only 6.27 face-action tracks.

\noindent
\underline{\bf A2D dataset}: This dataset \cite{xu2015can} contains 3782 YouTube videos (on average 7-10 seconds long) covering seven objects (ball, bird, car etc.) performing one of nine actions (fly, jump, roll etc.). The heterogeneous concepts considered are objects and actions. This dataset provides the bounding box annotations for every video label pair of object and action. Using the A2D dataset, we aim to analyze the track localization performance on weakly labeled videos as well as the track classification accuracy on a held-out test dataset. 

We use the method proposed in \cite{oneata2014spatio} to generate spatio-temporal object track proposals. For computational purpose, we consider only 10 tracks per video and use the Imagenet pretrained VGG CNN-M network \cite{Chatfield14} to generate object feature representation. We extract convolutional layer conv-4 and conv-5 features for each track image followed by PCA and video-level Fisher vector encoding. In  this dataset, the corresponding action tracks are kept similar to the object tracks (proposals) and the action features are extracted using the same approach as used for the Casablanca dataset.


\subsection{Baselines}
We compare WSC-SIIBP to several state-of-the-art approaches using the same features.
\begin{enumerate}[leftmargin=*,noitemsep]
\item {\bf WS-DC~\cite{bojanowski2013finding}}: This approach uses similar weak constraints as in \eqref{eq1}-\eqref{eq3}, but in a discriminative setup where the constraints are incorporated in a biconvex optimization framework.
\item {\bf WS-SIBP~\cite{shi2014weakly}}: This is a weakly supervised stacked IBP model which does not consider integrative framework for heterogeneous data and only enforces constraints equivalent to \eqref{eq4}-\eqref{eq5}. For each spatio-temporal track, the features extracted for heterogeneous concepts are concatenated while using this approach.
\item {\bf WS-S / WS-A}: This is similar to WS-SIBP except that instead of concatenating features from multiple concepts they are treated independently in two different IBP. WS-S is used to model only the person/object features and WS-A is used to model the action features.
\item {\bf WS-SIIBP}: This model integrates WS-SIBP with heterogeneous concepts. 
\item {\bf WSC-SIBP}: This model is similar to WS-SIBP, but unlike WS-SIBP, it additionally enforces the location constraints obtained from weak labels.
\end{enumerate}

\noindent
{\bf Implementation details:} For each dataset, the Fisher encoded features are PCA reduced to an appropriate dimension, $D^{e}$. We select the best feature length and other algorithm specific hyper-parameters for each algorithm using cross-validation on a small set of input videos. For the IBP based models, the cross-validation range for hyper-parameters are $K_{max} := K_a + K_s : 10 : K_a + K_s + 100$, $\alpha := 3K_{max}: 10 : 4K_{max}$ and $C := 0 : 0.5 : 5$. For all IBP based models, the parameters $D^{e}$, $\alpha$, $K_{max}$ and $C$ are set as 32, 100, 30 and 0.5 respectively for the Casablanca dataset and as 128, 160, 50 and 5 respectively for the A2D dataset. For WS-DC, $D^{e}$ is set as 1024.

\vspace{-0.5cm}
\begin{figure*}
\centering
\subfloat[]{
  \includegraphics[width=0.33\textwidth]{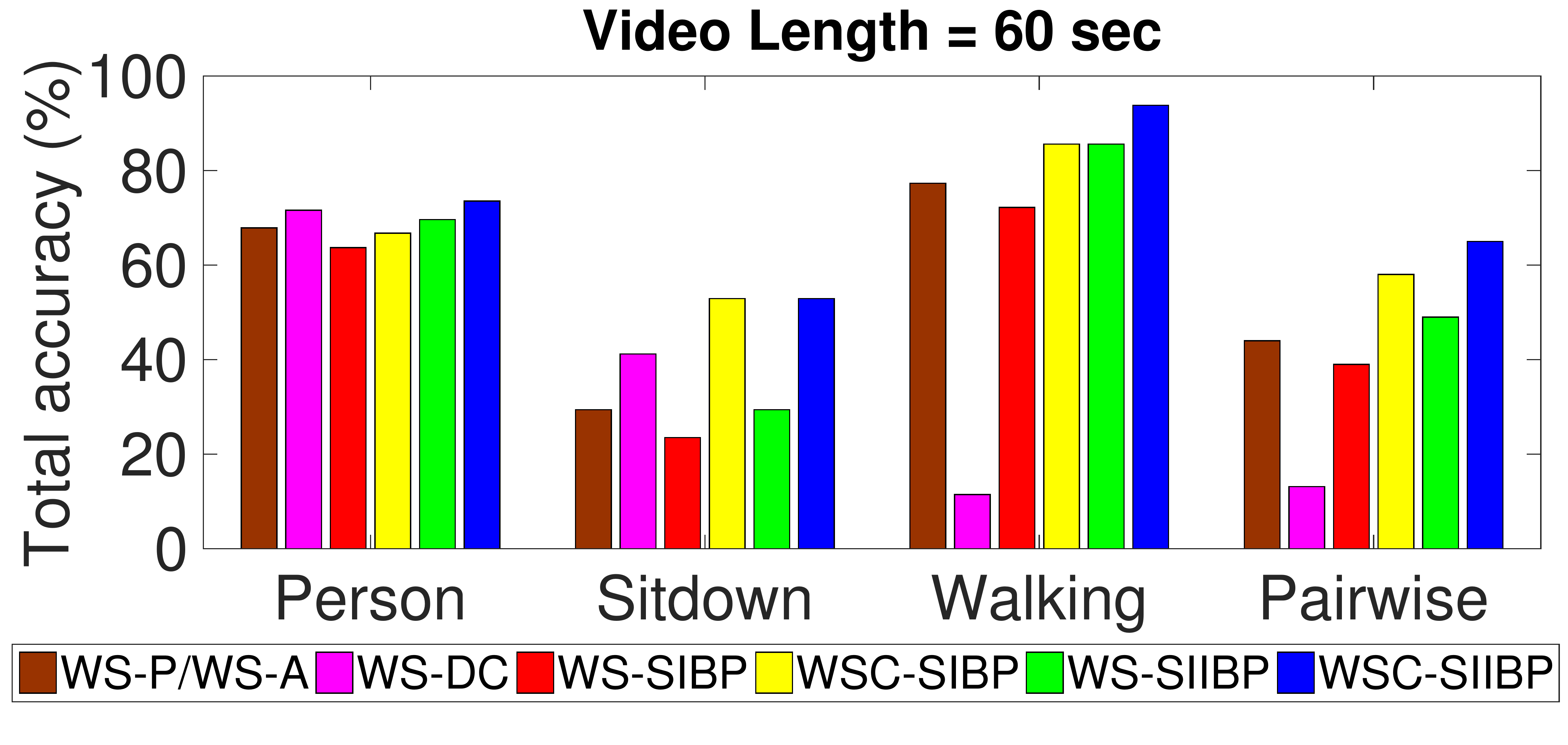}
  \label{cas_60}
}
\subfloat[]{
  \includegraphics[width=0.33\textwidth]{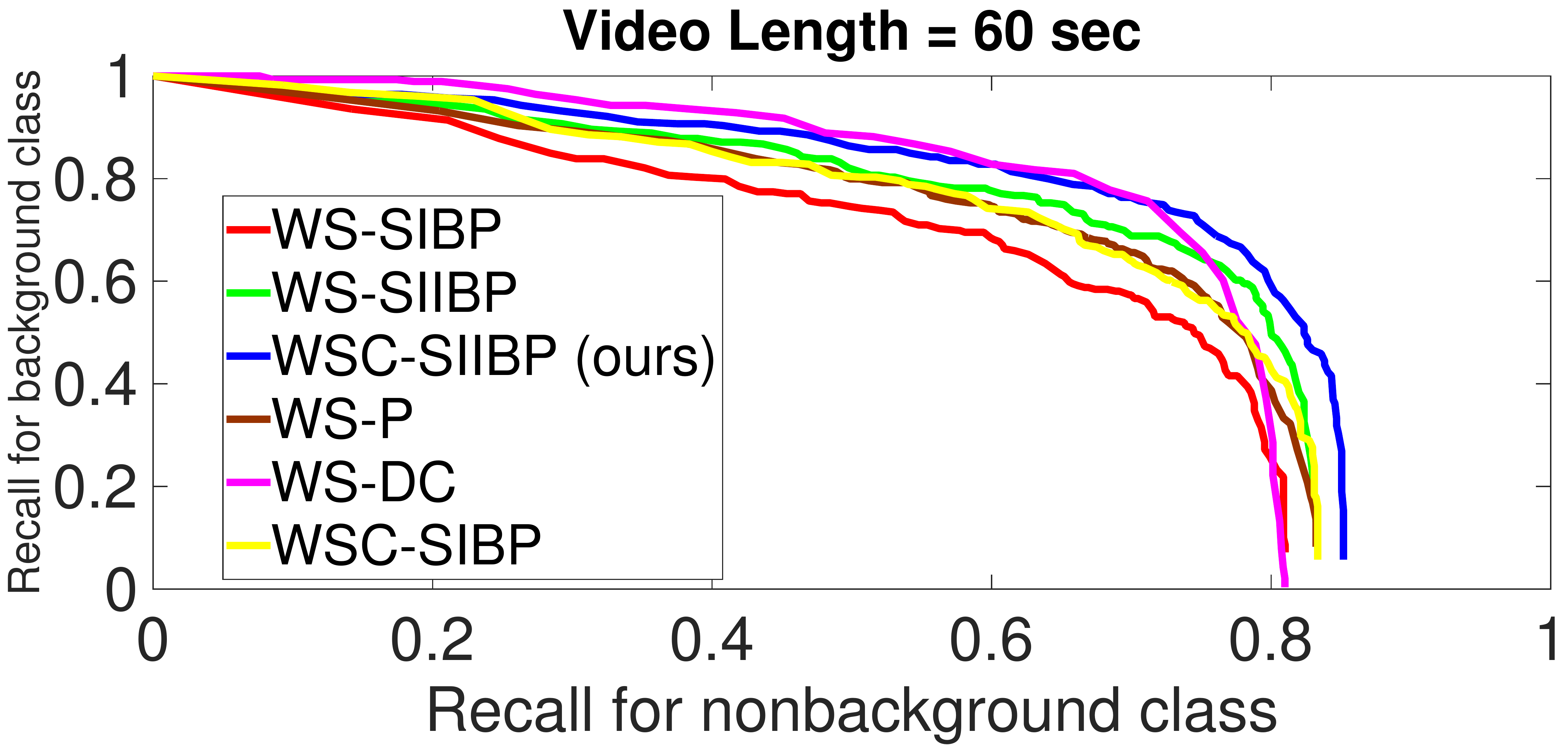}
  \label{cas_60_roc_1}
}
\subfloat[]{
  \includegraphics[width=0.33\textwidth]{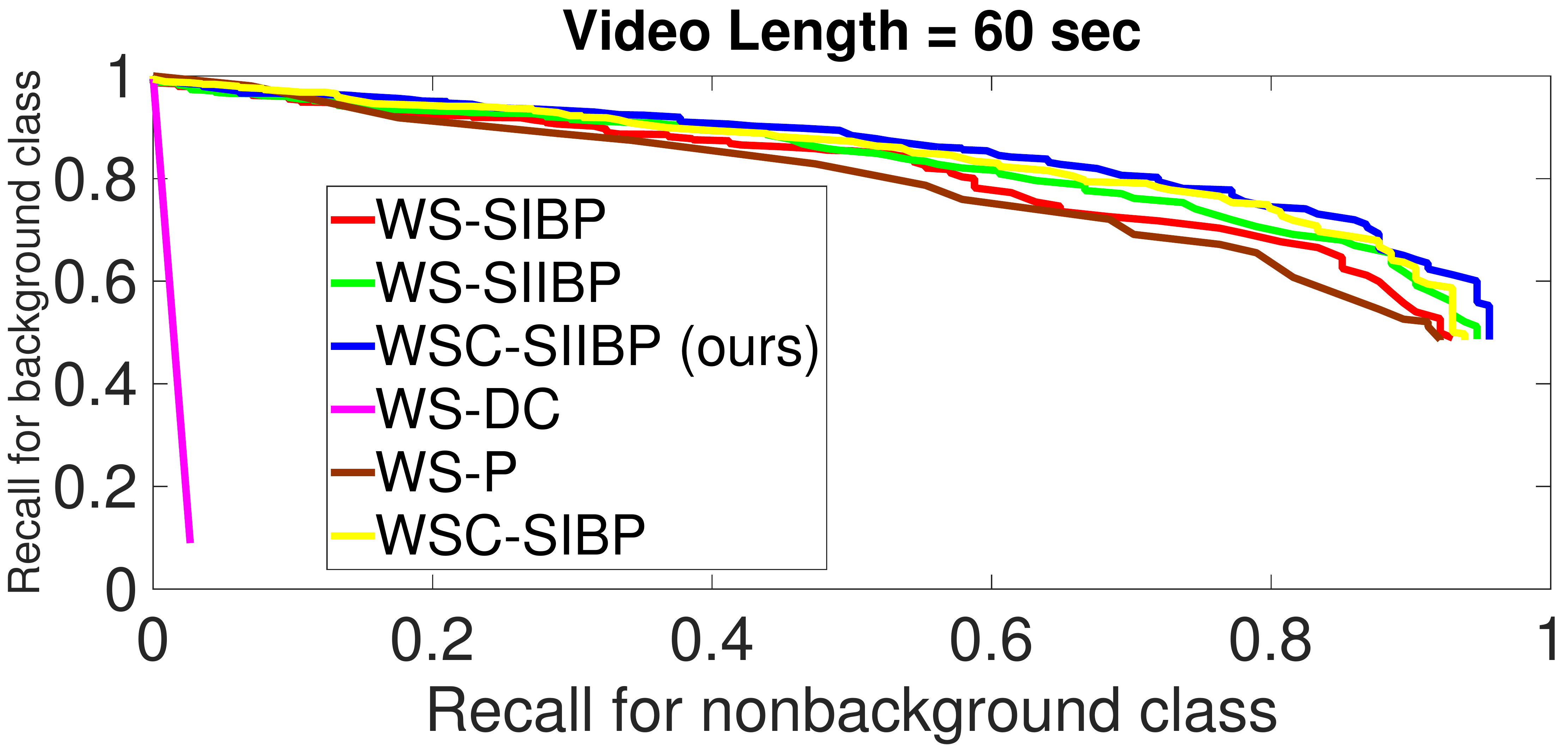}
  \label{cas_60_roc_2}
}\\
\vspace{-4mm}
\subfloat[]{
  \includegraphics[width=0.33\textwidth]{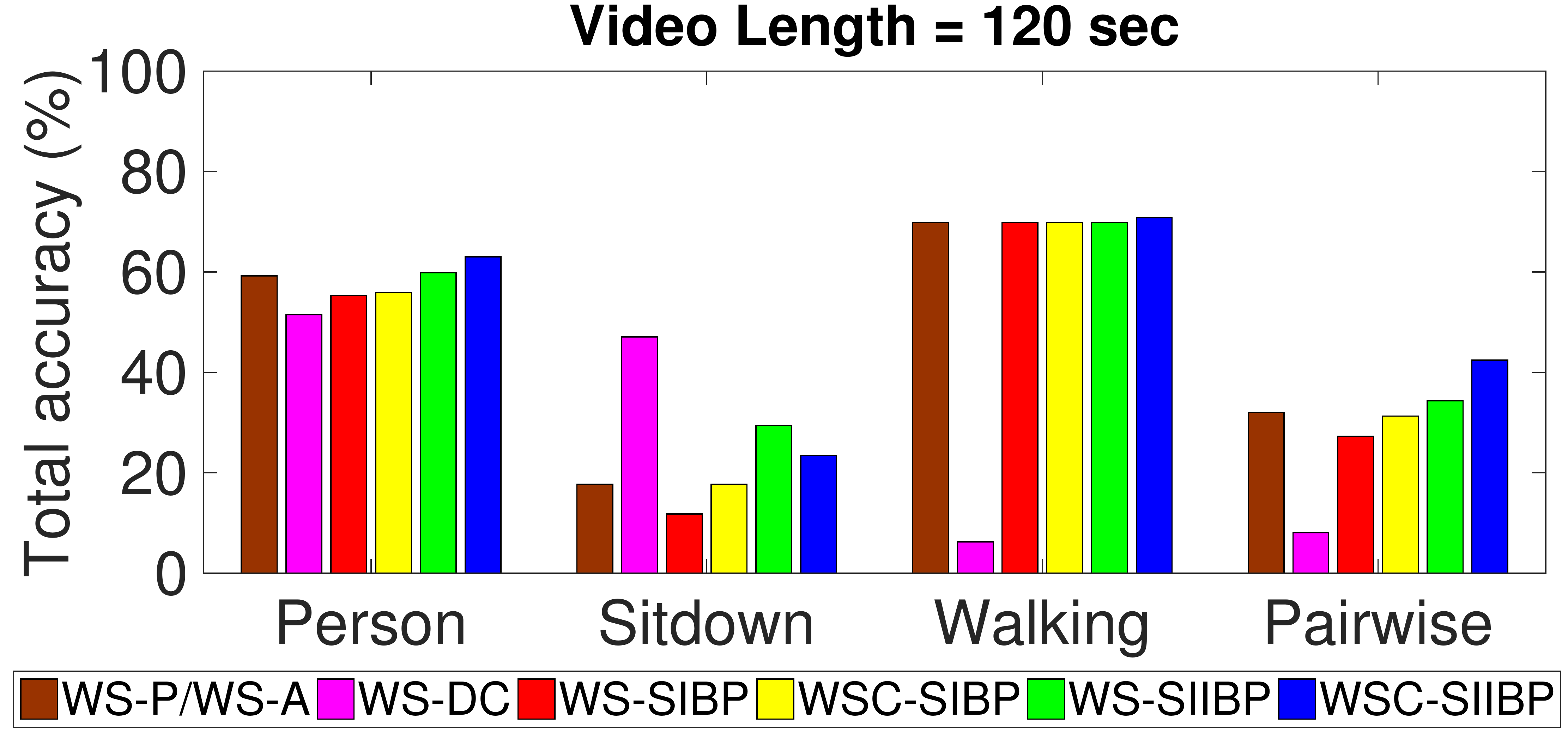}
  \label{cas_120}
}
\subfloat[]{
  \includegraphics[width=0.33\textwidth]{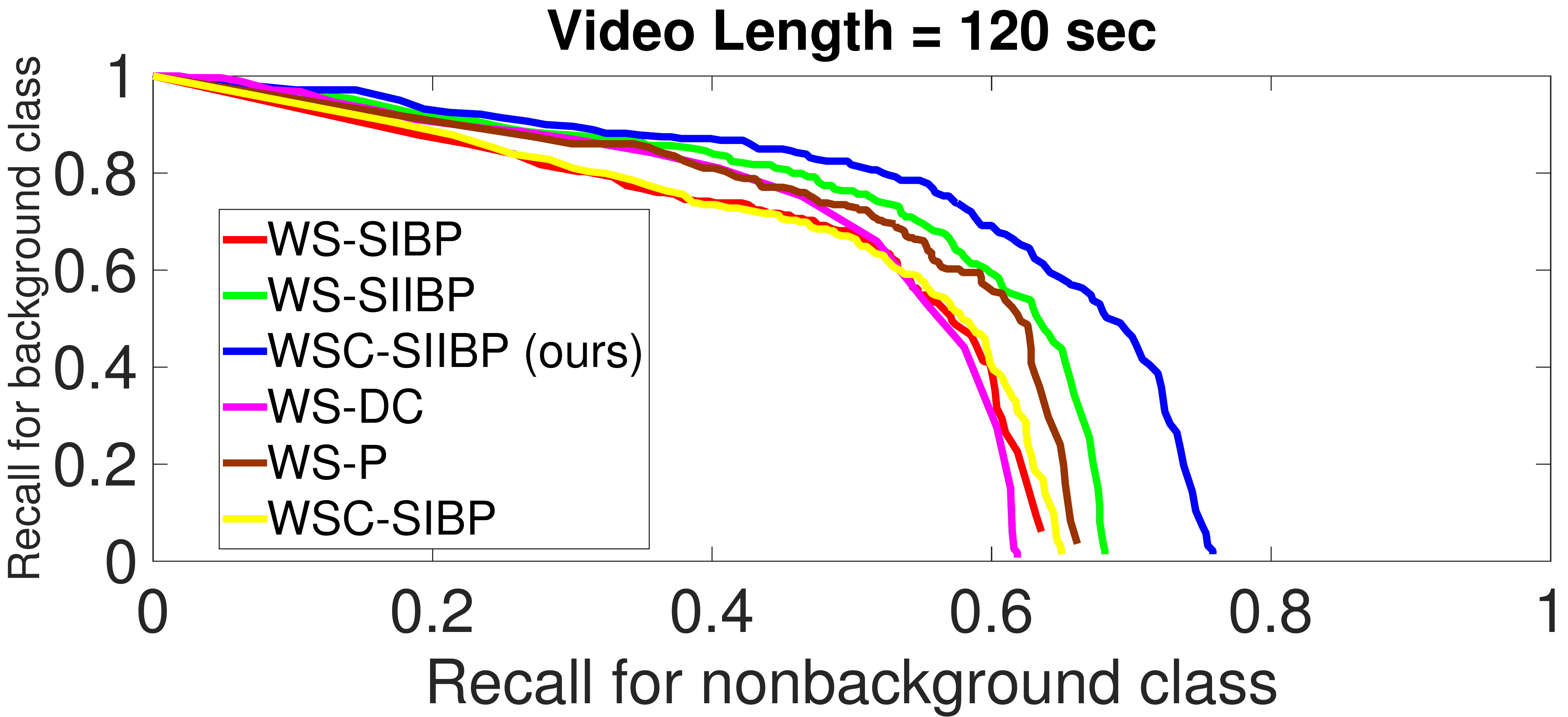}
  \label{cas_120_roc_1}
}
\subfloat[]{
  \includegraphics[width=0.33\textwidth]{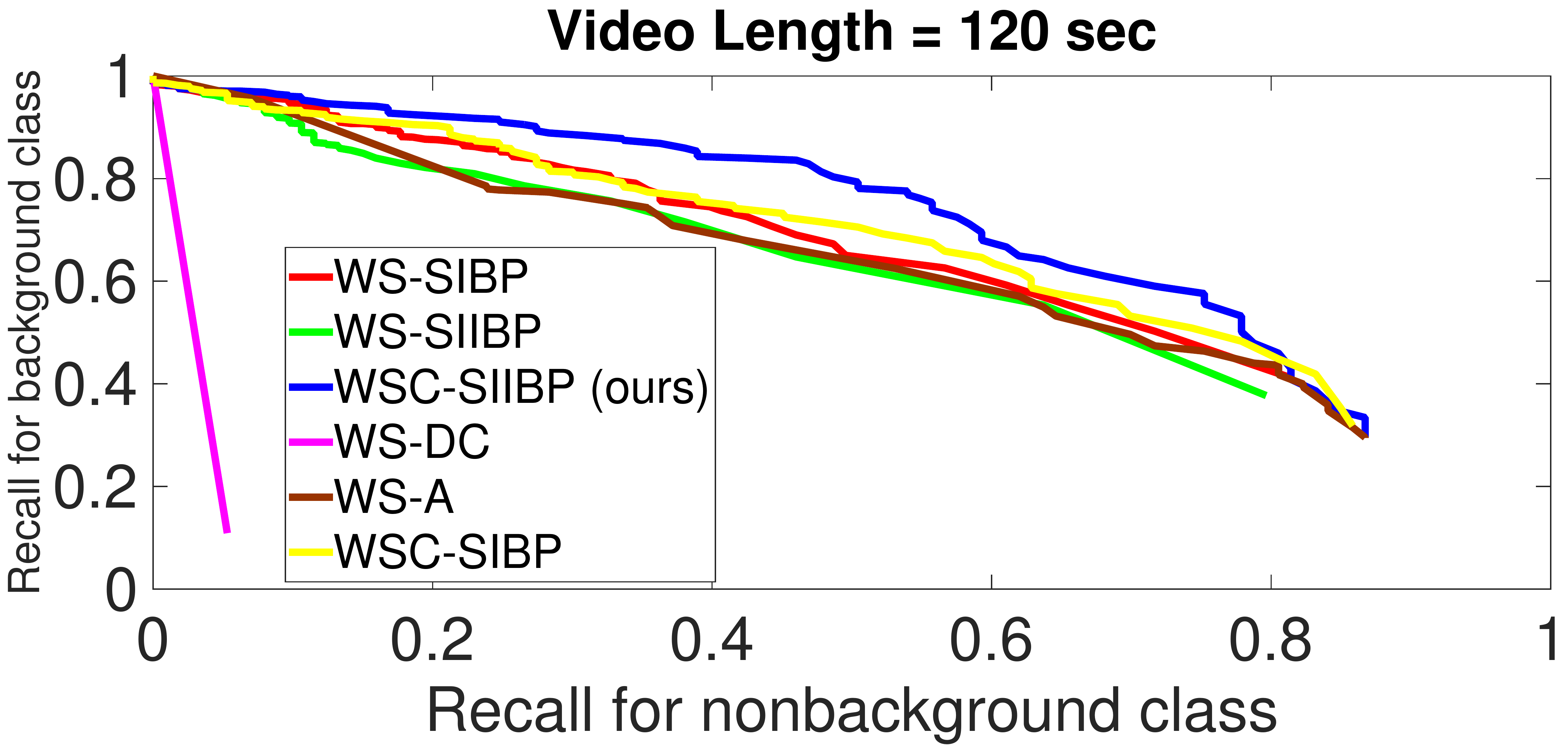}
  \label{cas_120_roc_2}
}\\
\vspace{-4mm}
\subfloat[]{
  \includegraphics[width=0.33\textwidth]{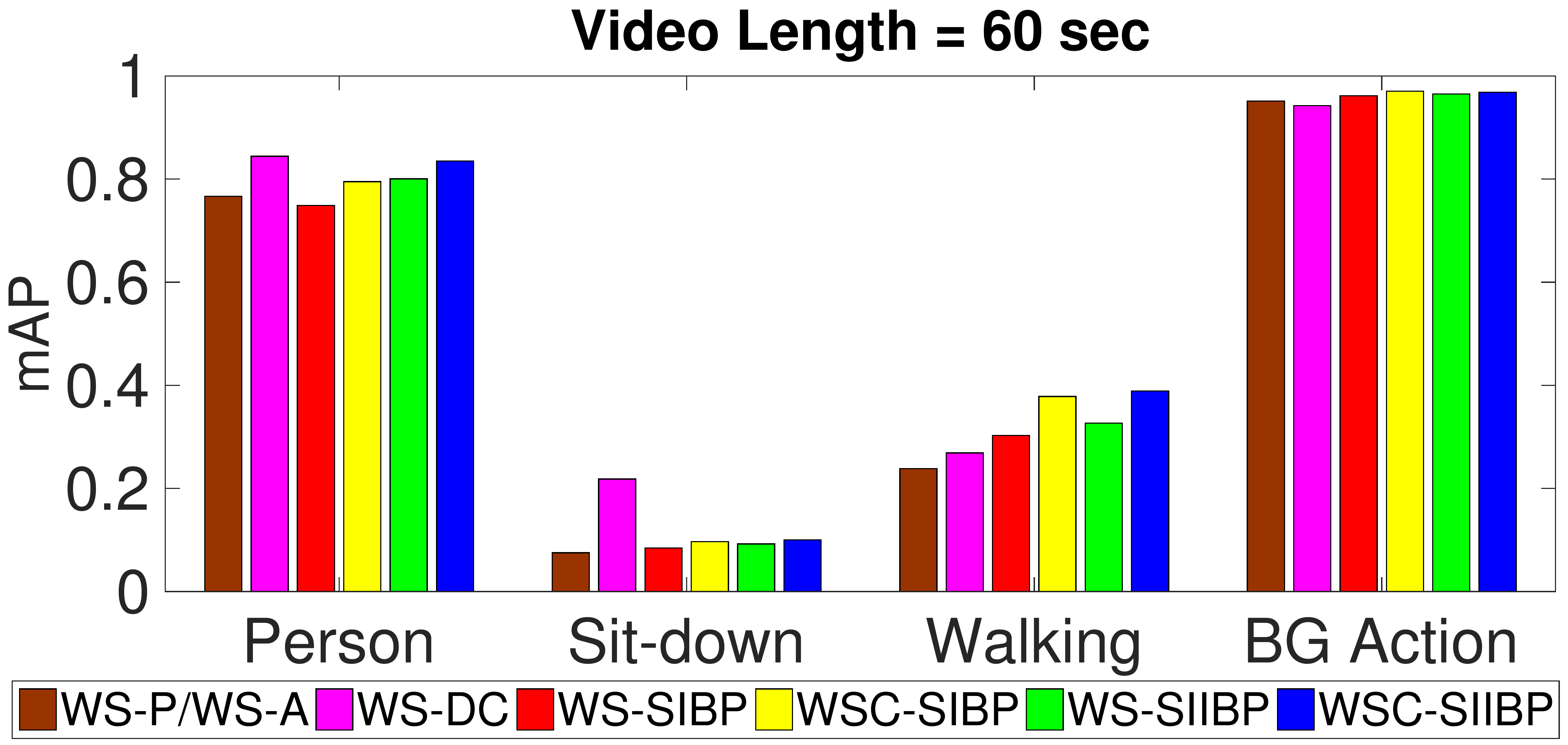}
  \label{map_60}
}
\subfloat[]{
  \includegraphics[width=0.33\textwidth]{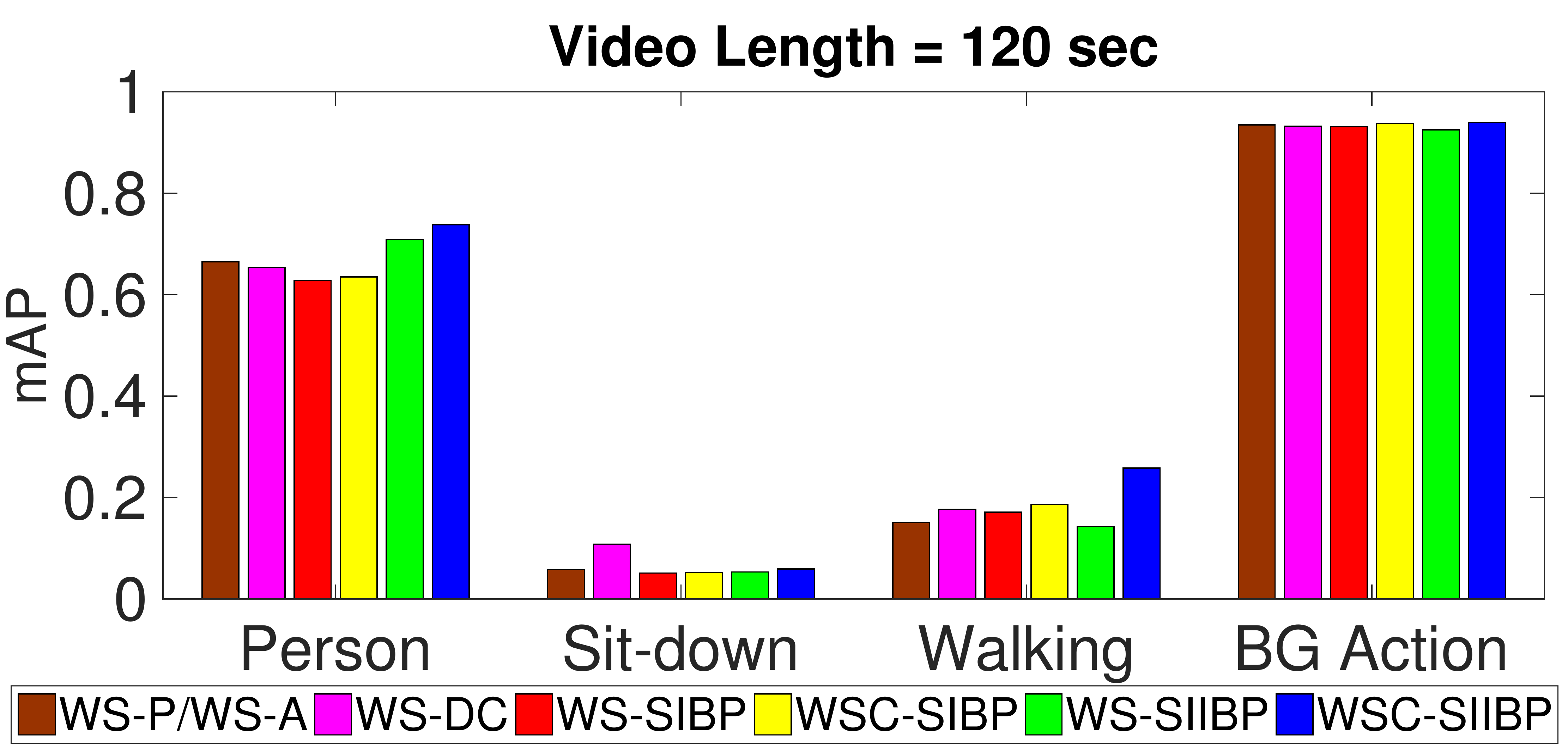}
  \label{map_120}
}
\subfloat[]{
  \includegraphics[width=0.33\textwidth]{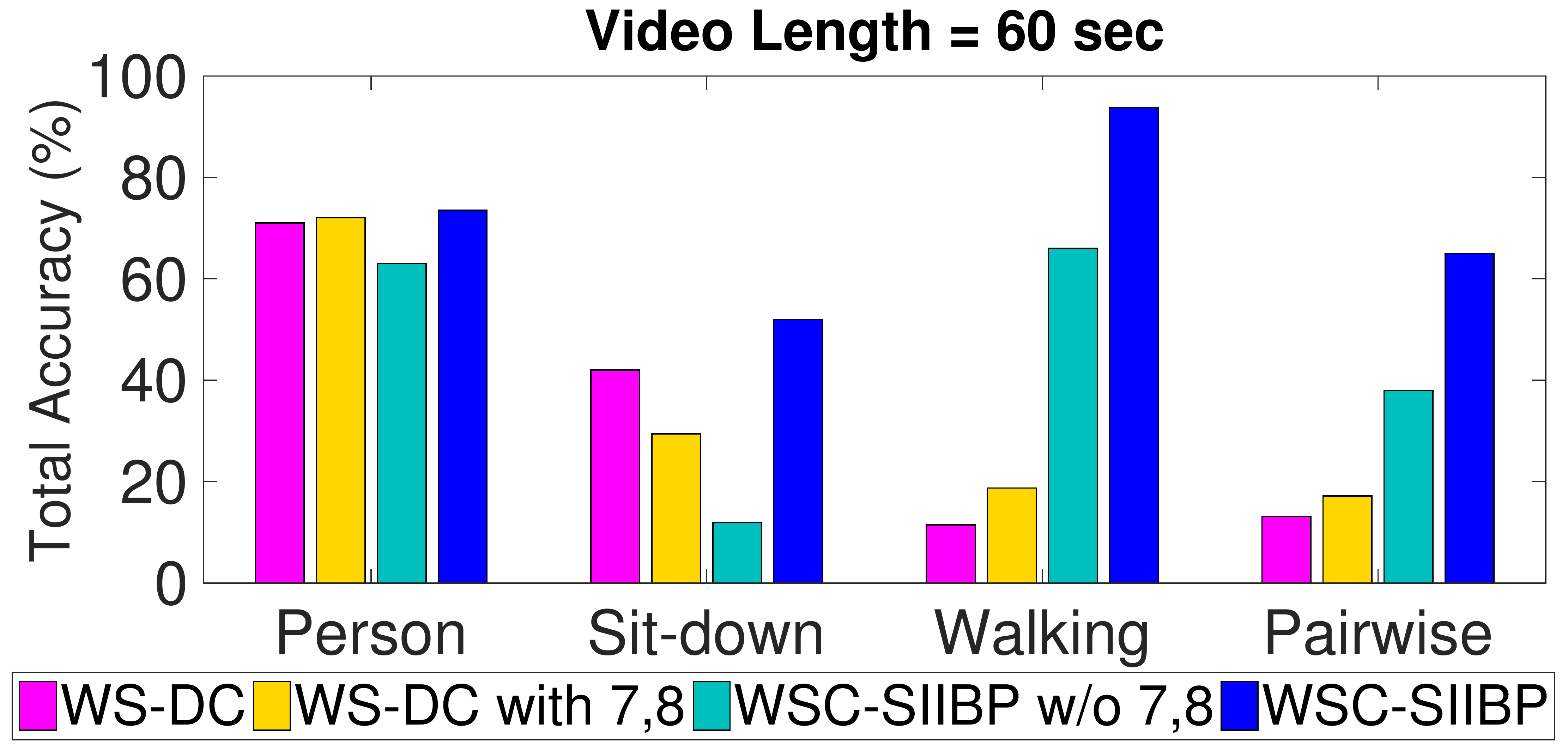}
  \label{cs_60}
}
\caption{Comparison of results for the Casablanca movie dataset. (a) Classification accuracy for 60 sec. segments. (b) Recall for background vs non-background class (60 sec., person). (c)  Recall for background vs non-background (60 sec., action). (d) Classification accuracy for 120 sec. segments. (e) Recall for background vs non-background class (120 sec., person). (f) Recall for background vs non-background (120 sec., action). (g),(h) Mean Average Precision for 60, 120 sec. segments. (i) Classification accuracy obtained with and without constraints \eqref{eq4} and \eqref{eq5}}
\label{Fig:experimental_result}
\vspace{-1cm}
\end{figure*}

\subsection{Results on Casablanca}
\label{sec:res_casa}
The track-level classification performance is compared in Figure \ref{Fig:experimental_result}. From Figures \ref{cas_60} and \ref{cas_120}, it can be seen that WSC-SIIBP significantly outperforms other methods for person and action classification in almost all of the scenarios. For instance, in the 120 second video segments, person classification improves by 4\% (relative improvement is 7\%) compared to the most competitive approach WS-SIIBP. We also compare pairwise label accuracy to gain insight into the importance of the constraints in eq \eqref{eq1}-\eqref{eq3}. For any given track with non-background person and action label, the classification is assumed to be correct only if both person and action labels are correctly assigned. Even in this scenario WSC-SIIBP performs 8.1\% better (24\% relative improvement) than the most competitive baseline. Since we combine the heterogeneous concepts along with location constraints in an integrated framework, WSC-SIIBP outperforms all other baselines. The weak results of WS-DC in pairwise classification, though surprising, can be attributed to their action classification results which are significantly biased towards one particular action `sitdown' (figure \ref{cas_120}, note that WS-DC performs very poorly in `walking' classification). Indeed, it should be noted that nearly 40\% and 89\% of person and action labels respectively belong to the background class. Thus, for fair evaluation of both background and non-background classes,  we also plot the recall of background class against the recall of nonbackground classes for person and action classification in Figure \ref{cas_60_roc_1}, \ref{cas_60_roc_2}, \ref{cas_120_roc_1}, \ref{cas_120_roc_2}. These curves were obtained by simultaneously computing recall for background and non-background classes at a range of threshold values on score, $\nu$. The mean average precision (mAP) of WSC-SIIBP along with all other baselines are plotted in Figure \ref{map_60} and \ref{map_120}. The mAP values also clearly demonstrate the effectiveness of the proposed approach. From the performance of WS-SIIBP (integrative concepts, no constraints) and WSC-SIBP (no integrative concepts, constraints) (Figure \ref{cas_60} and \ref{cas_120}), it is clear that the improvement in performance in the WSC-SIIBP can be attributed to both addition of integrative concepts and the location constraints.


\noindent
{\bf Effect of constraints \eqref{eq4}, \eqref{eq5}:} We note that, regardless of other differences, every weakly supervised IBP model considered here enforces constraints \eqref{eq4}, \eqref{eq5}. However, these constraints are not part of original WS-DC. To make a fair comparison between WS-DC and WSC-SIIBP, we analyze the effect of these constraints in Figure \ref{cs_60}. Although, these additional constraints improve WS-DC performance, they do not supersede the performance of WSC-SIIBP. Further we observe that these constraints have improved the performance of all the weakly supervised IBP models.

\subsection{Results on A2D}
\begin{figure*}[!tp]
\centering
\subfloat[]{
  \includegraphics[width=0.33\textwidth]{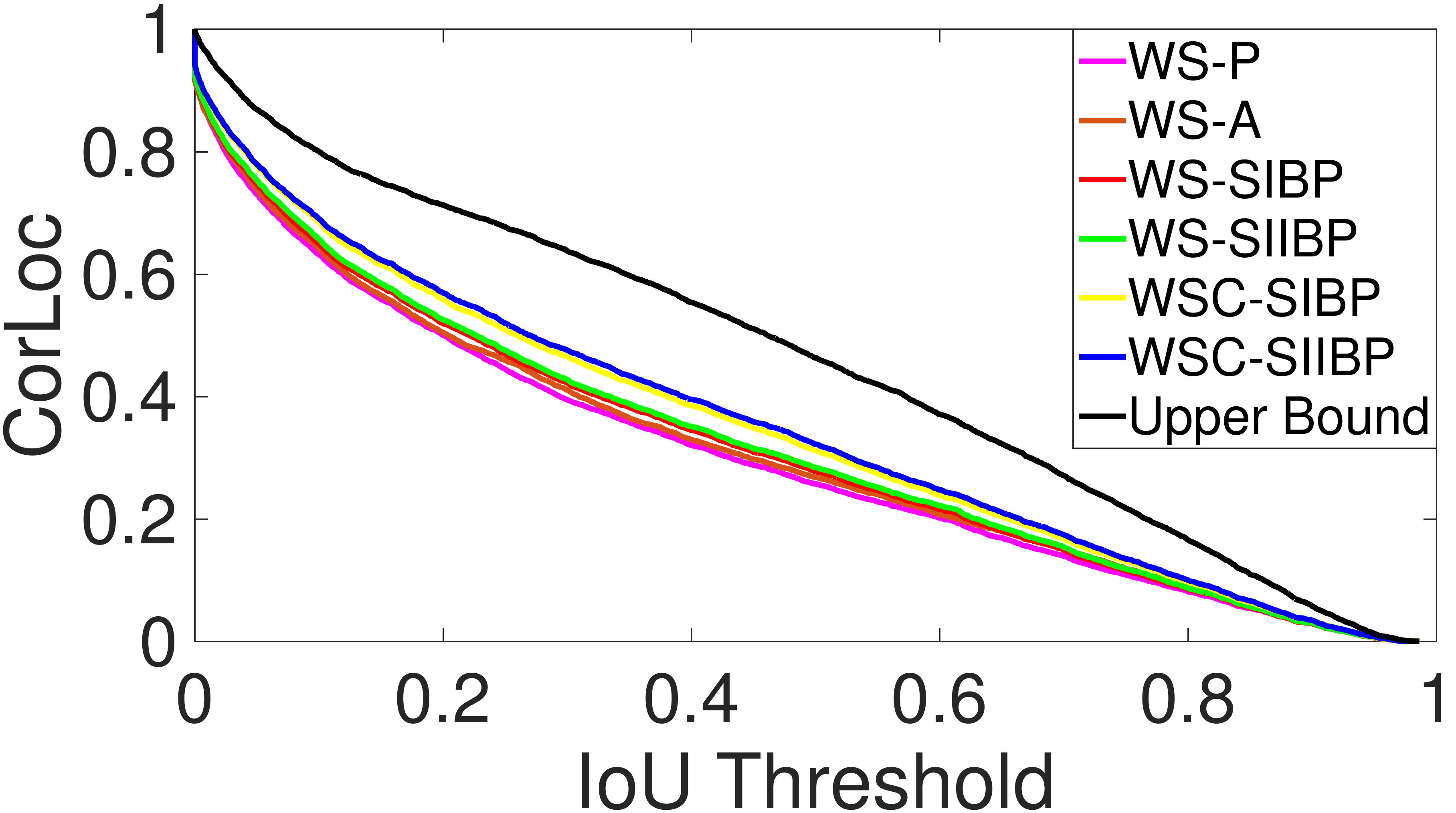}
  \label{Fig:iou_thresh}
}
\subfloat[]{
  \includegraphics[width=0.33\textwidth]{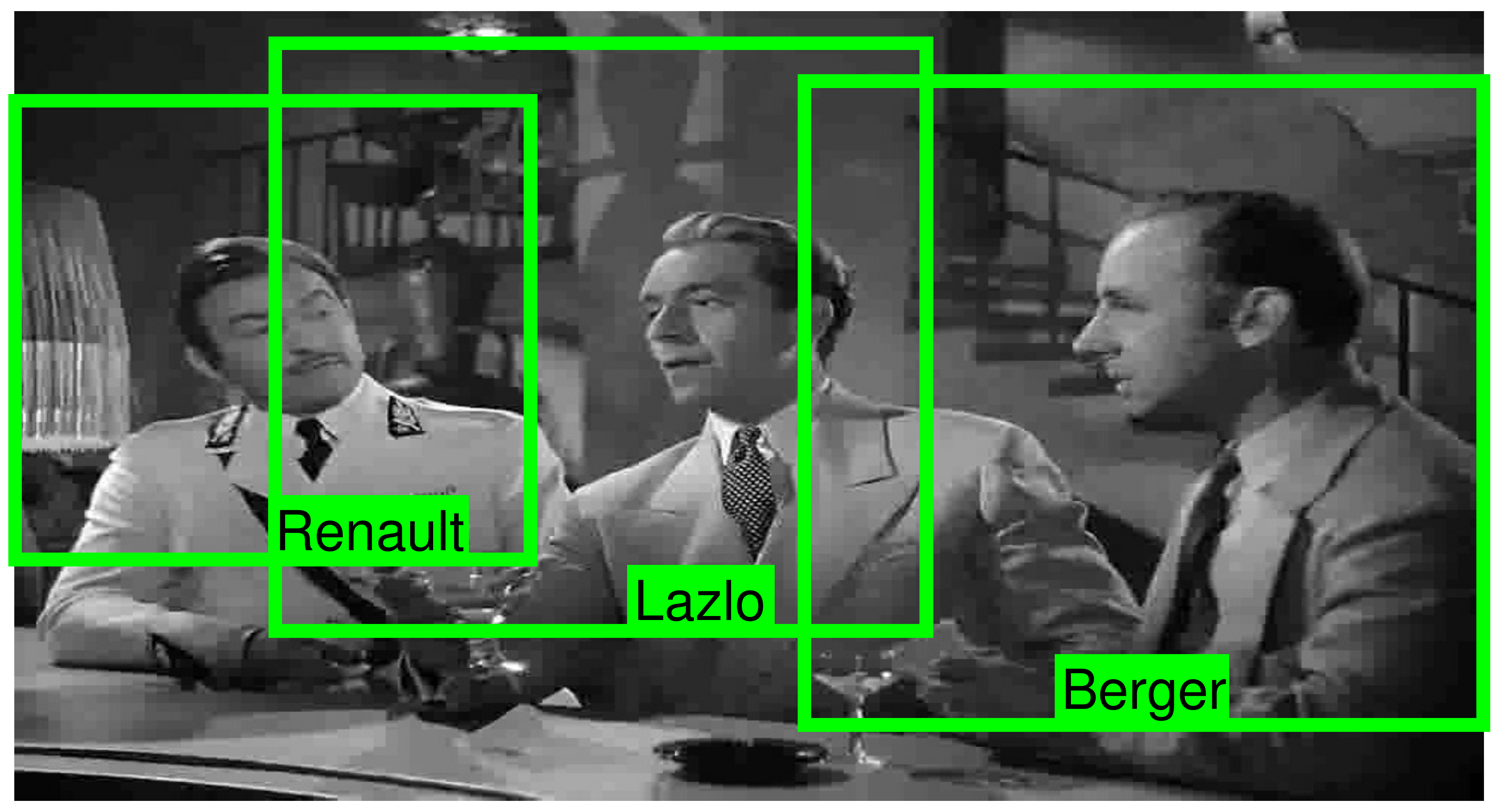}
  \label{qual_1}
}
\subfloat[]{
  \includegraphics[width=0.33\textwidth]{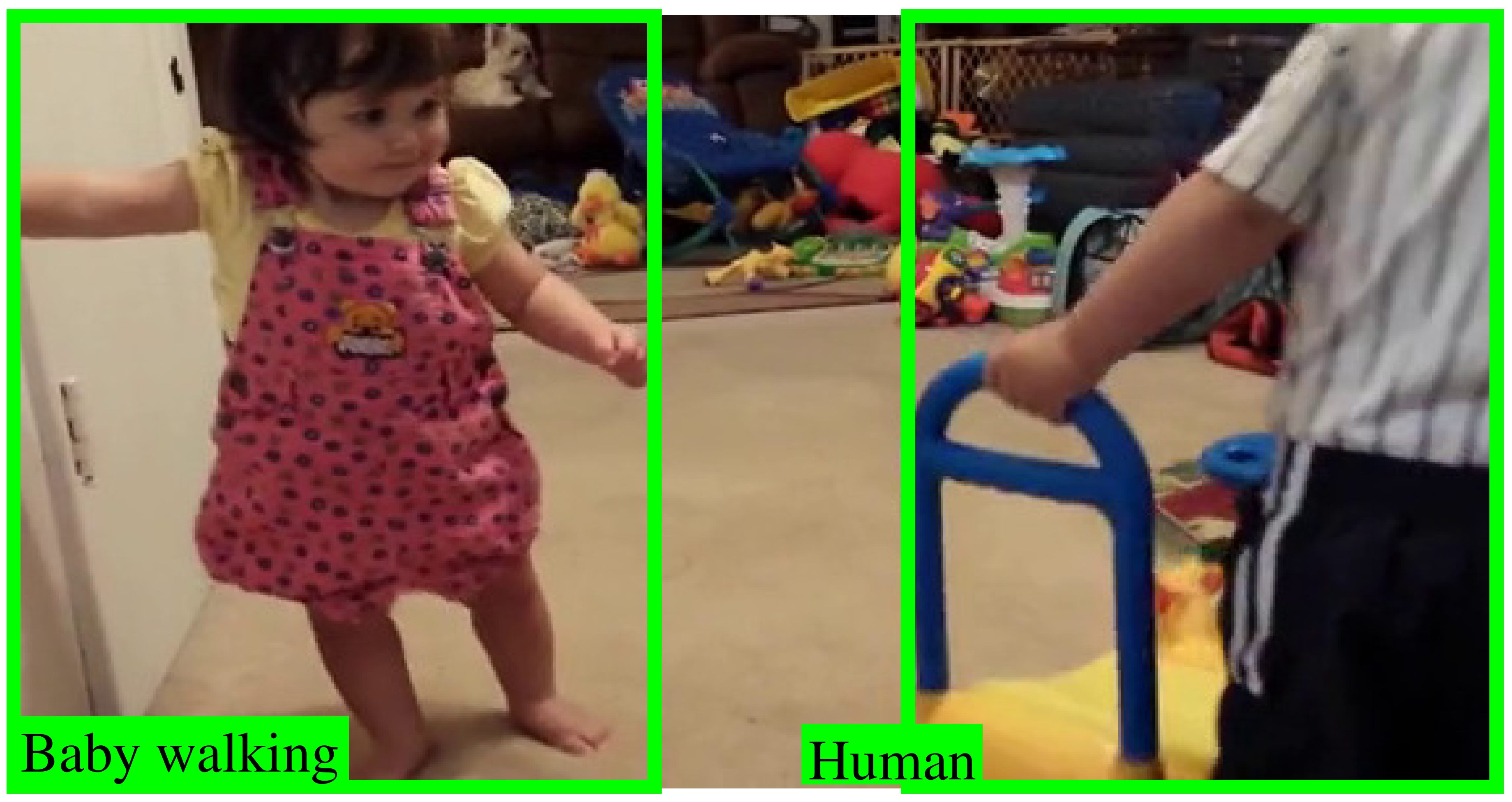}
  \label{qual_2}
}
\caption{(a) Correct localization accuracy at various IOU thresholds. (b) and (c) Qualitative results: green boxes show the concept localization using our proposed approach.}
\label{Fig:experimental_result_2}
\end{figure*}

First, we evaluate localization performance on the full A2D dataset. We experiment with 37,820 tracks extracted from 3,782 videos with around 5000 weak labels. For every given object-action label pair our algorithm selects the best track from the corresponding video using the approach outlined in (section \ref{sec:proposed_inference}). The localization accuracy is measured by calculating the average IoU (Intersection over Union) of the selected track (3-D bounding box) with the ground truth bounding box. The class-wise IoU accuracy and the mean IoU accuracy for all classes are tabulated in Table \ref{tb1} and \ref{tb2} respectively. In this task also WSC-SIIBP leads to a relative improvement of 9\% above the next best baseline. We also evaluate how accurately the extracted object proposals match with the ground truth bounding boxes to estimate an upper bound on the localization accuracy (referred as Upper Bound in Table \ref{tb1} and \ref{tb2}). In this case, the track maximizing the average IoU with the ground truth annotation is selected and the corresponding IoU is reported. We plot the correct localization accuracy with varying IoU thresholds in figure \ref{Fig:iou_thresh}, which also shows the effectiveness of the proposed approach. Figure \ref{qual_1}-\ref{qual_2} shows some qualitative localization results using the proposed approach on a few track images.
\newline
\noindent
\underline{\bf Test Inference}: We evaluate the classification performance on held-out test samples using the same train/test partition as in \cite{xu2015can}. We consider two setups for the evaluation, (a) using video-level labels for the test samples and (b) free annotation where no test video labels are provided. The proposed approach is compared with GT-SVM, which is a fully supervised linear SVM that uses ground truth bounding boxes and their corresponding strong labels during training. The results are tabulated in Table \ref{tb3}. Note that the performance of WSC-SIIBP is close to that of the fully supervised setup. 

\begin{table*}[!tp]
\parbox{\linewidth}{
\centering
\small
\begin{tabular}{| c ||p{0.59cm}|p{0.55cm}|p{0.55cm}|p{0.55cm}|p{0.55cm}|p{0.55cm}|p{0.55cm}|p{0.65cm}|p{0.62cm}|p{0.55cm}|p{0.55cm}|p{0.59cm}|p{0.55cm}|p{0.55cm}|p{0.55cm}|}
\hline
& \rot{adult} & \rot{baby} & \rot{ball} & \rot{bird} & \rot{car} & \rot{cat} & \rot{dog} & \rot{climb} & \rot{crawl} & \rot{eat} & \rot{fly} & \rot{jump} & \rot{roll} & \rot{run} & \rot{walk} \\ \hline
WSC-SIIBP &28.4&43.6&9.8&37.8&37.4&40.8&42.0&37.5&47.6&46.1&24.5&29.4&50.9&25.6&37.2 \\ \hline
Upper Bound &39.9&53.9&16.4&48.2&48.7&52.8&51.4&50.0&59.2&57.2&33.9&41.0&59.1&38.1&47.9\\ \hline
\end{tabular} 
\caption{{\small Per class mean IoU on A2D dataset.}}
\label{tb1}
}
\\
\parbox{\linewidth}{
\centering
\small
\begin{tabular}{|c||@{\hspace{0.2em}}c@{\hspace{0.2em}}|@{\hspace{0.2em}}c@{\hspace{0.2em}}|@{\hspace{0.2em}}c@{\hspace{0.2em}}|@{\hspace{0.2em}}c@{\hspace{0.2em}}|@{\hspace{0.2em}}c@{\hspace{0.2em}}|@{\hspace{0.2em}}c@{\hspace{0.2em}}|@{\hspace{0.2em}}c@{\hspace{0.2em}}|@{\hspace{0.2em}}c@{\hspace{0.2em}}|}
\hline
 & Random&WS-P&WS-A&WS-SIBP&WS-SIIBP&WSC-SIBP&WSC-SIIBP& Upper Bound \\ \hline
IoU &25.5&29.7&30.43 & 31.1& 31.55 & 31.69 & {\bf 34.38} & 45.05 \\ \hline
\end{tabular} 
\caption{{\small Average IoU comparison with other approaches on A2D dataset.}}
\label{tb2}
}
\hfill
\parbox{\linewidth}{
\centering
\small
\begin{tabular}{|L{3cm}||p{1cm}|p{1cm}|p{1cm}|p{1cm}|}
\hline
& \multicolumn{2}{c|}{WSC-SIIBP} & \multicolumn{2}{c|}{GT-SVM} \\ \hline
Setup & Obj & Act & Obj & Act \\ \hline
Using video Labels &94.77&90.68&98.20&94.92 \\ \hline
Free Annotation &76.62&64.77&85.18&73.26  \\ \hline
\end{tabular} 
\caption{{\small mAP classification test accuracy on A2D dataset.}}
\label{tb3}
}
\vspace{-1cm}
\end{table*}


\section{Conclusion}
We developed a Bayesian non-parametric approach that integrates Indian Buffet Process with heterogeneous concepts and spatio-temporal location constraints arising from weak labels. 
We perform experimental results on two recent datasets containing heterogeneous concepts such as persons, objects and actions and show that our approach outperforms the best state of the art method. In future work, we will extend the WSC-SIIBP model to additionally localize audio concepts from speech input and develop an end-to-end deep neural network for joint feature learning and Bayesian inference.

\renewcommand{\theequation}{S\arabic{equation}}
\section{Appendix}
\subsection{Expectation Constraints}
In an Bayesian framework, the effective constraints for equations \eqref{eq1}-\eqref{eq5} are defined as an expectation \cite{zhu2014bayesian,ganchev2010posterior} of the original constraints and can be rewritten as,
\begin{align} 
&\forall i \in 1\dots M, \nonumber\\
& \sum_{j=1}^{N_i} \mathbb{E}_{\tilde{q}}\left[z_{js}^{(i)} z_{ja}^{(i)}\right] \geq 1 - \xi_{(s,a)}^{(i)}, \quad \forall (s,a)\in \Gamma^{(i)} \label{eq6} \\[-0.5em]
& \sum_{j=1}^{N_i} \mathbb{E}_{\tilde{q}}\left[z_{js}^{(i)}\right] \geq 1 - \xi_{(s,\emptyset)}^{(i)}, \quad \forall (s,\emptyset)\in \Gamma^{(i)} \label{eq7}\\[-0.5em]
& \sum_{j=1}^{N_i}  \mathbb{E}_{\tilde{q}}\left[z_{ja}^{(i)}\right] \geq 1 - \xi_{(\emptyset,a)}^{(i)}, \quad \forall (\emptyset,a)\in \Gamma^{(i)} \label{eq8} \\[-0.5em]
&\forall i \in 1\dots M \mbox{ and }\forall j \in 1\dots N_i, \nonumber\\
& \mathbb{E}_{\tilde{q}}\left[z_{js}^{(i)}\right] = 0, \quad \hspace{-3.5mm}\text{if } \nexists (s,\emptyset)\in\Gamma^{(i)} \mbox{ and } \nexists (s,a) \in \Gamma^{(i)}, \forall a \in \mathcal{A}\label{eq9}\\[-0.5em]
& \mathbb{E}_{\tilde{q}}\left[z_{ja}^{(i)}\right] = 0, \quad \hspace{-3.5mm}\text{if } \nexists (\emptyset,a)\in\Gamma^{(i)} \mbox{ and } \nexists  (s,a) \in \Gamma^{(i)},  \forall s \in \mathcal{S} \label{eq10}
\vspace{-3mm}
\end{align}
where the expectation is taken w.r.t.\ the posterior distribution in \eqref{eq3.3a}. From \eqref{eq3.3a} one may note that through $\pi_a^{(i)}$, the samples of $z_{ja}^{(i)}$ depends on the previously sampled latent coefficients such as $z_{js}^{(i)}$. This complicates the applicability of constraint in equation \eqref{eq6}. However due to the independency assumption, the search space over the family of tractable posterior distribution in \eqref{eq12} simplifies the constraint in equation \eqref{eq6}-\eqref{eq10} to, 
\begin{align}
&\forall i \in 1\dots M, \nonumber\\
& \sum_{j=1}^{N_i} \nu_{js}^{(i)} \nu_{ja}^{(i)} \geq 1 - \xi_{(s,a)}^{(i)}, \quad \forall (s,a)\in \Gamma^{(i)} \label{eq13}\\[-0.5em]
& \sum_{j=1}^{N_i} \nu_{js}^{(i)} \geq 1 - \xi_{(s,\emptyset)}^{(i)}, \quad \forall (s,\emptyset)\in \Gamma^{(i)} \label{eq14}\\[-0.5em]
& \sum_{j=1}^{N_i}  \nu_{ja}^{(i)} \geq 1 - \xi_{(\emptyset,a)}^{(i)}, \quad \forall (\emptyset,a)\in \Gamma^{(i)} \label{eq15}\\[-0.5em]
&\forall i \in 1\dots M \mbox{ and }\forall j \in 1\dots N_i, \nonumber\\
& \nu_{js}^{(i)} = 0, \text{if } \nexists (s,\emptyset)\in\Gamma^{(i)} \mbox{ and } (s,a) \in \Gamma^{(i)}, \forall a \in \mathcal{A} \label{eq16}\\[-0.3em]
& \nu_{ja}^{(i)} = 0, \text{if } \nexists (\emptyset,a)\in\Gamma^{(i)} \mbox{ and } (s,a) \in \Gamma^{(i)},  \forall s \in \mathcal{S} \label{eq17}
\end{align}

\subsection{Derivation of Posterior Update Equations}
Now, note that the constraints in \eqref{eq13}-\eqref{eq15} can be rewritten as hinge loss function and added as part of the objective function in equation \eqref{eq11}.  Hence the final formulation is given by,

\begin{equation}
{\small
\begin{aligned}
\label{eq18}
\min_{\substack{\mathbf{\nu}^{(i)}, \mathbf{\tau}^{(i)}, \\ \mathbf{\Phi}_k^*, \sigma_{k*}^2}} &\quad KL\left(\tilde{w}(\mathbf{Y}) || \tilde{\Psi}(\mathbf{Y}|\mathbf{\Theta})\right) 
 - \sum_{i=1}^M \sum_{j=1}^{N_i} \int \left(\sum_{e\in\{s,a\}}\log p\left(\mathbf{X^e}^{(i)}_j | \mathbf{Y}, \mathbf{\Theta}\right) \right) \tilde{w}(\mathbf{Y}) d\mathbf{Y}   \\
 & + C \sum_{i=1}^M \left(\sum_{\substack{J \in \Gamma^{(i)} \\ J = (s,a)}} \max\left(0,1-\sum_{j=1}^{N_i}  \nu_{js}^{(i)}\nu_{ja}^{(i)}\right)
+ \sum_{\substack{J \in \Gamma^{(i)} \\ J = (s,\emptyset)}} \max\left(0,1-\sum_{j=1}^{N_i}  \nu_{js}^{(i)}\right) \right.\\
& \left.+ \sum_{\substack{J \in \Gamma^{(i)} \\ J = (\emptyset,a)}} \max\left(0,1-\sum_{j=1}^{N_i} \nu_{ja}^{(i)}\right)\right) \\
s.t. & \quad \forall i \in 1\dots M, \text{ and } \forall j \in 1\dots N_i, \quad \eqref{eq16}, \eqref{eq17}
\end{aligned}
}
\end{equation}

\noindent The objective function in eq. \eqref{eq18} can be rewritten as,
\begin{align}
L(\mathbf{\nu}^{(i)}, \mathbf{\tau}^{(i)}, \mathbf{\Phi}_k^*, \sigma_{k*}^2) = \mathcal{L} - \sum_{i=1}^M\sum_{j=1}^{N_i} \left( L_{ij} - C  \sum_{k=1}^{K_a + K_s}H_{jk}^{(i)}\right) \label{eq19}
\end{align}
where $\mathcal{L}$ represent KL-divergence term, $L_{ij}$ denote the likelihood term and $H_{jk}$ is the term corresponding to hinge loss function for $\nu_{jk}^{(i)}$. Expanding $L_{ij}$, we get,
\begin{align}
L_{ij} &\triangleq \mathbb{E}_{\tilde{w}}\left[\log p(\mathbf{X^s}^{(i)}_j | \mathbf{Y}, \mathbf{\Theta}) + \log p(\mathbf{X^a}^{(i)}_j | \mathbf{Y}, \mathbf{\Theta})\right] \label{ref1} \\
&= - \frac{{\mathbf{x^s}_j^{(i)}}^T\mathbf{x^s}_j^{(i)} -2 \mathbb{E}_{\tilde{w}}[\mathbf{z}_{j.}^{(i)}\mathbf{A^s}]\mathbf{x^s}_j^{(i)} + \mathbb{E}_{\tilde{w}}[\mathbf{z}_{j.}^{(i)}\mathbf{U^s}{\mathbf{z}_{j.}^{(i)}}^T]}{2\sigma_{ns}^2} - \frac{D^s\log(2\pi\sigma_{ns}^2)}{2} \nonumber \\
& \quad - \frac{{\mathbf{x^a}_j^{(i)}}^T\mathbf{x^a}_j^{(i)} -2 \mathbb{E}_{\tilde{w}}[\mathbf{z}_{j.}^{(i)}\mathbf{A^a}]\mathbf{x^a}_j^{(i)} + \mathbb{E}_{\tilde{w}}[\mathbf{z}_{j.}^{(i)}\mathbf{U^a}{\mathbf{z}_{j.}^{(i)}}^T]}{2\sigma_{na}^2} - \frac{D^a\log(2\pi\sigma_{na}^2)}{2}
\end{align}
where $\mathbf{U^*} = \mathbb{E}_{\tilde{w}}[\mathbf{A^*}\mathbf{A^*}^T]$ is $K_{max} \times K_{max}$ matrix, $U_{jk}^* = \mathbf{\Phi^*}_{j.}\mathbf{\Phi^*}_{k.}^T$; $\mathbb{E}_{\tilde{w}}[\mathbf{z}_{j.}^{(i)}\mathbf{A^*}]\mathbf{x^*}_j^{(i)} = \left(\sum_k \nu_{jk}^{(i)} \mathbf{\Phi^*}_{k.}\right)\mathbf{x^*}_j^{(i)} $; and 
\begin{align}
\mathbb{E}_{\tilde{w}}[\mathbf{z}_{n.}^{(i)}\mathbf{U^*}{\mathbf{z}_{n.}^{(i)}}^T] = 2 \sum_{j<k} \nu_{nj}^{(i)}\nu_{nk}^{(i)} U_{jk}^* + \sum_k \nu_{nk}^{(i)} \left(D^*\sigma_{k*}^2 + \mathbf{\Phi^*}_{k.}\mathbf{\Phi^*}_{k.}^T\right)
\end{align}
For KL-divergence term, we get $\text{KL}\left(\tilde{w}(\mathbf{Y}) || \tilde{\Psi}(\mathbf{Y}|\mathbf{\Theta})\right) = \text{KL}\left(\tilde{w}(\mathbf{v}) || \tilde{\Psi}(\mathbf{v}|\mathbf{\Theta})\right) + \text{KL}\left(\tilde{w}(\mathbf{Z}) || \tilde{\Psi}(\mathbf{Z}|\mathbf{\Theta})\right) + \text{KL}\left(\tilde{w}(\mathbf{A^s}) || \tilde{\Psi}(\mathbf{A^s}|\mathbf{\Theta})\right) + \text{KL}\left(\tilde{w}(\mathbf{A^a}) || \tilde{\Psi}(\mathbf{A^a}|\mathbf{\Theta})\right) $, where the individual terms are,
\begin{align}
& \text{KL}\left(\tilde{w}(\mathbf{v}) || \tilde{\Psi}(\mathbf{v}|\mathbf{\Theta})\right) = \nonumber \\
& \sum_{i=1}^M \left(\sum_{k=1}^{K_{max}} \left((\tau_{k1}^{(i)} - \alpha)(\Psi(\tau_{k1}^{(i)}) - \Psi(\tau_{k1}^{(i)} + \tau_{k2}^{(i)}))    + (\tau_{k2}^{(i)} - 1)(\Psi(\tau_{k2}^{(i)}) - \Psi(\tau_{k1}^{(i)} + \tau_{k2}^{(i)})) \right.\right.\nonumber \\
& \left.\left.- \log \frac{\Gamma(\tau_{k1}^{(i)}) \Gamma(\tau_{k2}^{(i)})}{\Gamma(\tau_{k1}^{(i)} + \tau_{k2}^{(i)})}\right) - K_{max}\log \alpha \right)\\
& \text{KL}\left(\tilde{w}(\mathbf{Z}) || \tilde{\Psi}(\mathbf{Z}|\mathbf{\Theta})\right) = \nonumber \\
& \sum_{i=1}^M \left( \sum_{j=1}^{N_i}\sum_{k=1}^{K_{max}} \left( -\nu_{jk}^{(i)} \sum_{j=1}^{K_{max}} (\Psi(\tau_{k1}^{(i)}) - \Psi(\tau_{k1}^{(i)} + \tau_{k2}^{(i)})) - (1-\nu_{jk}^{(i)}) \mathbb{E}_{\tilde{w}}[\log(1 - \prod_{j=1}^k v^{(i)})] \right.\right. \nonumber \\
& \left.\left.+\nu_{jk}^{(i)} \log \nu_{jk}^{(i)} + (1 - \nu_{jk}^{(i)})\log(1 - \nu_{jk}^{(i)})\right)\right) \\
&\text{KL}\left(\tilde{w}(\mathbf{A^*}) || \tilde{\Psi}(\mathbf{A^*}|\mathbf{\Theta})\right) = 
\sum_{k=1}^{K_{max}} \left(\frac{D^*\sigma_{k*}^2 + \mathbf{\Phi^*}_k\mathbf{\Phi^*}_k^T}{2\sigma_{A*}^2} - \frac{D^*\left(1 + \log\frac{\sigma_{k*}^2}{\sigma_{A*}^2}\right)}{2}\right)
\end{align}
where $\Psi(.)$ is the digamma function. As shown for original IBP in \cite{doshi2009variational}, the term $\mathbb{E}_{\tilde{w}}[\log(1 - \prod_{j=1}^k v^{(i)})]$ is approximated by its lower bound,
\begin{align}
\mathbb{E}_{\tilde{w}}[\log(1 - \prod_{j=1}^k v^{(i)})] &\geq \sum_{m=1}^k q_{km} \Psi(\tau_{m2}^{(i)}) + \sum_{m=1}^{k-1}\left(\sum_{n=m+1}^k q_{kn}\right)\Psi(\tau_{m1}^{(i)}) \nonumber \\
&- \sum_{m=1}^{k}\left(\sum_{n=m}^k q_{kn}\right)\Psi(\tau_{m1}^{(i)} + \tau_{m2}^{(i)}) + \mathcal{H}(q_{k.}) \label{ref2} \\
&= \mathcal{L}_k \nonumber
\end{align}
where the variational parameter $q_{k.} = (q_{k1}\dots q_{kk})$ is k-point probability mass function and $\mathcal{H}(q_{k.})$ denotes entropy of $q_{k.}$. The tightest upper bound is obtained by setting,
\begin{align*}
q_{km} = \frac{1}{Z_k} \exp\left(\Psi(\tau_{m2}^{(i)}) + \sum_{n=1}^{m-1}\Psi(\tau_{n1}^{(i)}) - \sum_{n=1}^m\Psi(\tau_{n1}^{(i)} + \tau_{n2}^{(i)})\right)
\end{align*} 
where $Z_k$ is the normalization factor to enable $q_{k.}$ to be a distribution. On replacing the term $\mathbb{E}_{\tilde{w}}[\log(1 - \prod_{j=1}^k v_j^{(i)})]$ with its lower bound $\mathcal{L}_k$, we have an upper bound for $\text{KL}\left(\tilde{w}(\mathbf{Y}) || \tilde{\Psi}(\mathbf{Y}|\mathbf{\Theta})\right)$.

On substituting equation \eqref{ref1}-\eqref{ref2} in \eqref{eq19}, the optimum value for parameters of mean-field variational approximate posterior distribution \eqref{eq12} are obtained by setting the derivative of \eqref{eq19} w.r.t. those parameters to zero and simultaneously solving for all parameters (using KKT conditions). We derive the following equations which are iteratively solved,
\begin{align}
\sigma_{ke}^2 &= \left(\frac{1}{\sigma_{Ae}^2} + \frac{1}{\sigma_{ne}^2}\sum_{i=1}^M \sum_{j=1}^{N_i}\nu_{jk}^{(i)}\right)^{-1}, \,\, \forall e \in \{s,a\} \label{eq20}\\
\mathbf{\Phi^e}_k &= \left(\frac{1}{\sigma_{ne}^2}\sum_{i=1}^M \sum_{j=1}^{N_i} \nu_{jk}^{(i)} \left(\mathbf{x^e}_j^{(i)} - \sum_{l:l\neq k}\nu_{jl}^{(i)}\mathbf{\Phi^e}_l\right)\right)\sigma_{ke}^2, \,\, \forall e \in \{s,a\} \label{eq21}\\
\tau_{k1}^{(i)} &= \alpha + \sum_{m=k}^{K_{max}}\sum_{j=1}^{N_i} \nu_{jm}^{(i)} + \sum_{m=k+1}^{K_{max}} \left(N_i - \sum_{j=1}^{N_i} \nu_{jm}^{(i)}\right)\left(\sum_{s = k+1}^m q_{ms}^{(i)}\right) \label{eq22}\\
\tau_{k2}^{(i)} &= 1 + \sum_{m=k}^{K_{max}} \left(N_i - \sum_{j=1}^{N_i} \nu_{jm}^{(i)}\right)q_{mk}^{(i)} \label{eq23}
\end{align}
Above equations are somewhat similar to those given by variational approximation on IBP \cite{doshi2009variational}. The update equation for $\nu$ differs completely and it is given by,
\begin{align}
\nu_{jk}^{(i)} &= \frac{L_k^{(i)}}{1+e^{-\zeta_{jk}^{(i)}}}  \label{eq24.2}\\
\zeta_{jk}^{(i)} &= \sum_{t=1}^k \left(\Psi(\tau_{t1}^{(i)}) - \Psi(\tau_{t1}^{(i)} + \tau_{t2}^{(i)})\right) - \mathcal{L}_k - \frac{1}{2\sigma_{ns}^2}\left(D^s\sigma_{ks}^2 + \mathbf{\Phi^s}_k\mathbf{\Phi^s}_k^T\right) \nonumber \\
& - \frac{1}{2\sigma_{na}^2}\left(D^a\sigma_{ka}^2 + \mathbf{\Phi^a}_k\mathbf{\Phi^a}_k^T\right) 
+ \frac{1}{\sigma_{ns}^2}  \mathbf{\Phi^s}_k\left(\mathbf{x}_j^{(i)} - \sum_{l\neq k} \nu_{jl}^{(i)}\mathbf{\Phi^s}_l\right)^T \nonumber \\
& + \frac{1}{\sigma_{na}^2}  \mathbf{\Phi^a}_k\left(\mathbf{x}_j^{(i)} - \sum_{l\neq k} \nu_{jl}^{(i)}\mathbf{\Phi^a}_l\right)^T + C\sum_{\substack{J\in \Gamma^{(i)} \\ J=(k,a)}} \mathbb{I}_{\left\lbrace\sum_{l=1}^{N_i}\nu_{lk}^{(i)}\nu_{la}^{(i)} < 1\right\rbrace} \nu_{ja}^{(i)}  \nonumber \\
&+ C\sum_{\substack{J\in \Gamma^{(i)} \\ J=(s,k)}} \mathbb{I}_{\left\lbrace\sum_{l=1}^{N_i}\nu_{ls}^{(i)}\nu_{lk}^{(i)} < 1\right\rbrace} \nu_{js}^{(i)} + C \mathbb{I}_{\left\lbrace\sum_{l=1}^{N_i}\nu_{lk}^{(i)} < 1, k \leq K_a + K_s\right\rbrace} \label{eq25.2}
\end{align}
where $L_k^{(i)}$ and $\mathbb{I}$ is an indicator variable. $L_k^{(i)}$ indicates whether an entity (action / subject) $k$ is part of $i^{th}$ video label set $\Gamma^{(i)}$ or not. This inturn enforces $\nu = 0$ for all $\nu$ satisfying eq. \eqref{eq16} and eq. \eqref{eq17}. 

The hyperparameter $\sigma_{n*}^2$ and $\sigma_{A*}^2$ can be set apriori or estimated from the data. The empirical estimation can be easily derived by maximizing the expected log-likelihood, which is similar to maximization step of EM algorithm. The closed form solution is given by,
\begin{align}
\sigma_{A*}^2 &= \frac{\sum_{k=1}^{K_{max}} D^*\sigma_{k*}^2 + \mathbf{\Phi^*}_k\mathbf{\Phi^*}_k^T}{K_{max}D^*} \label{eq26}\\
\sigma_{n*}^2 &= \frac{\sum_{i=1}^M\sum_{j=1}^{N_i} \left({\mathbf{x^*}_j^{(i)}}^T\mathbf{x^*}_j^{(i)} -2 \mathbb{E}_{\tilde{w}}[\mathbf{z}_{j.}^{(i)}\mathbf{A^*}]\mathbf{x^*}_j^{(i)} + \mathbb{E}_{\tilde{w}}[\mathbf{z}_{j.}^{(i)}\mathbf{U^*}{\mathbf{z}_{j.}^{(i)}}^T]\right)}{(\sum_{i=1}^M N_i) D} \label{eq27}
\end{align}

The final algorithm is summarized in Algorithm \ref{inference} in the paper.

\subsection{Additional Experimental Results}

In this section we share some of the results which provides additional insights into experiments.

\begin{figure}[!h]
\centering
\includegraphics[width=1\textwidth]{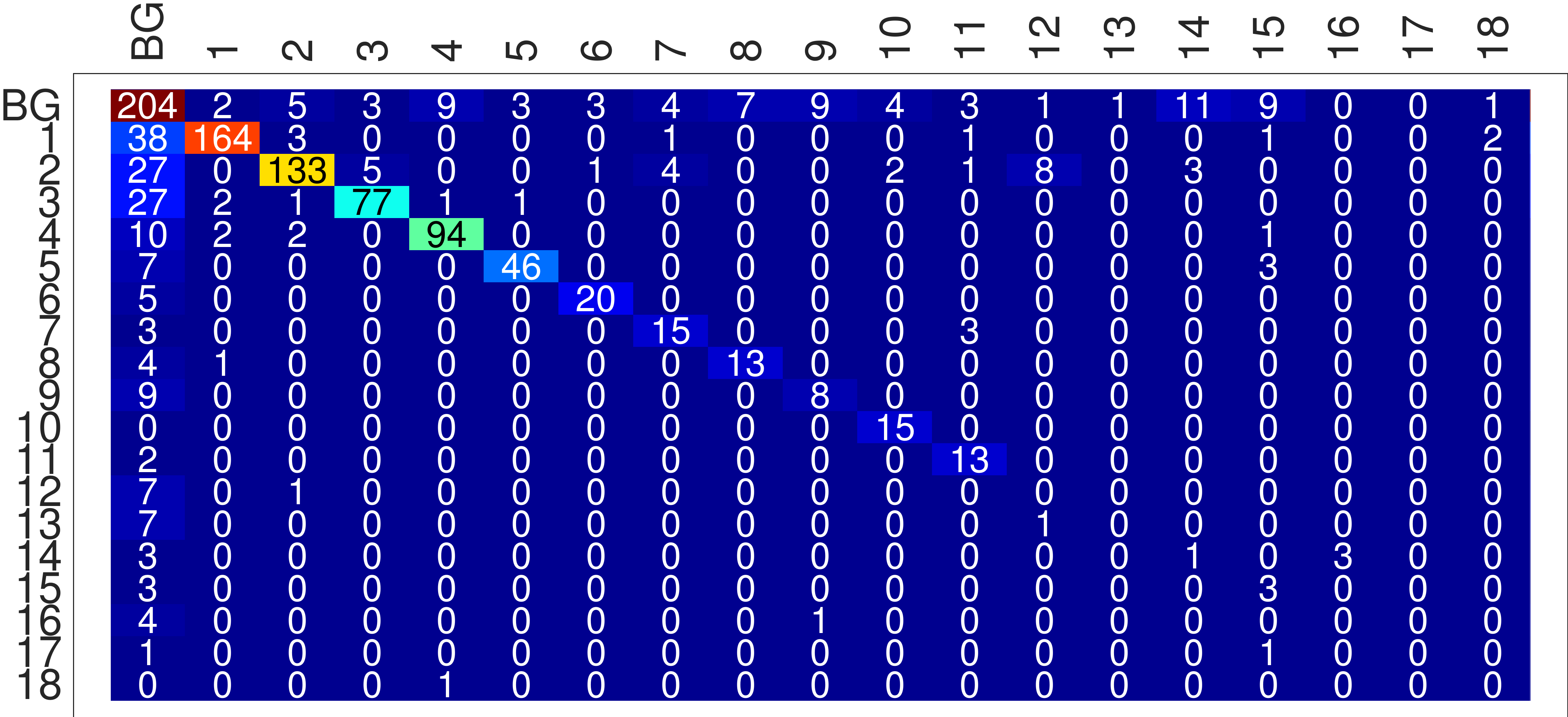}
\caption{Person class confusion matrix. BG denotes the background class which can represent any unknown face.}
\label{Fig1}
\end{figure}

{\bf Casablanca:} The person class confusion matrix is shown in Figure \ref{Fig1}. It exhibits that our approach learns each person appearance model with high accuracy and it can learn from as less as 15 weakly annotated samples.

\begin{figure}[!h]
\centering
\includegraphics[width=0.49\textwidth]{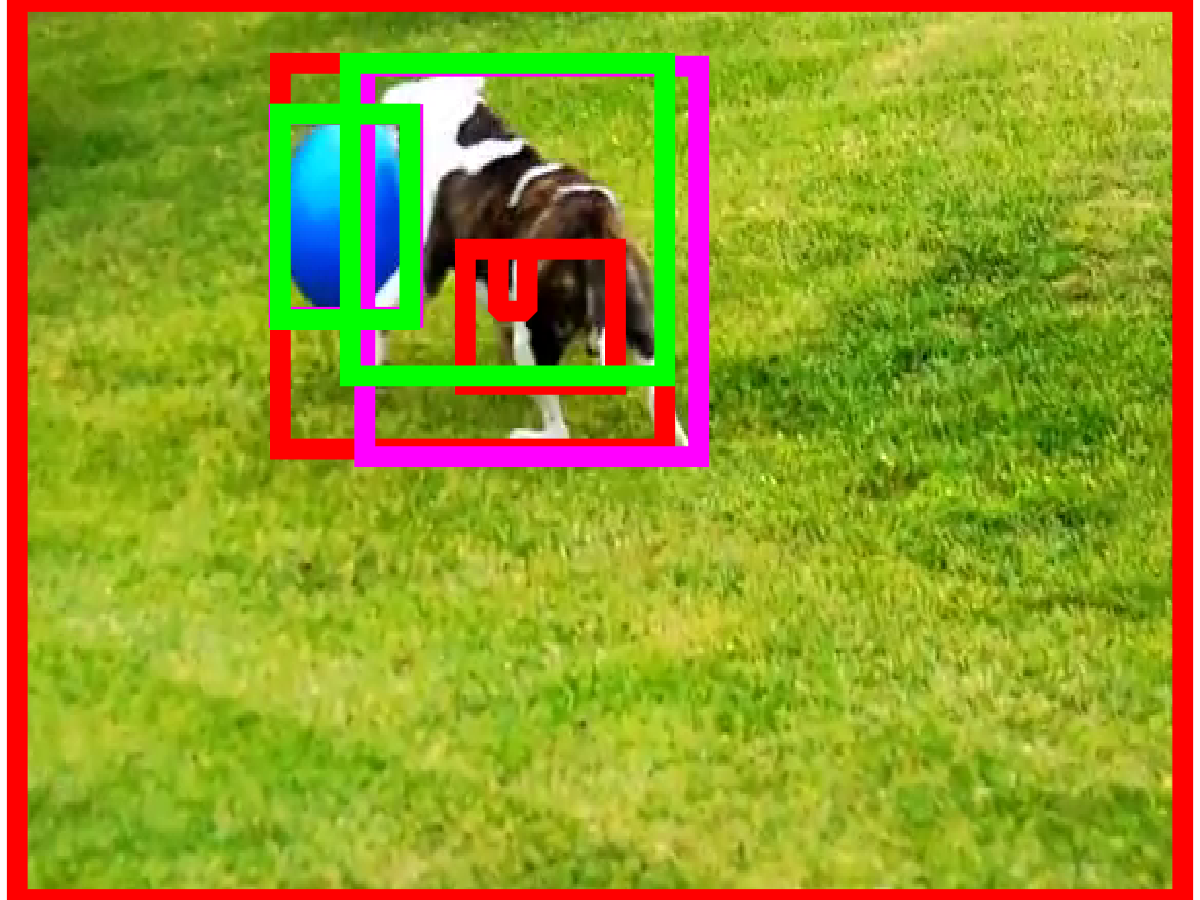}
\includegraphics[width=0.49\textwidth]{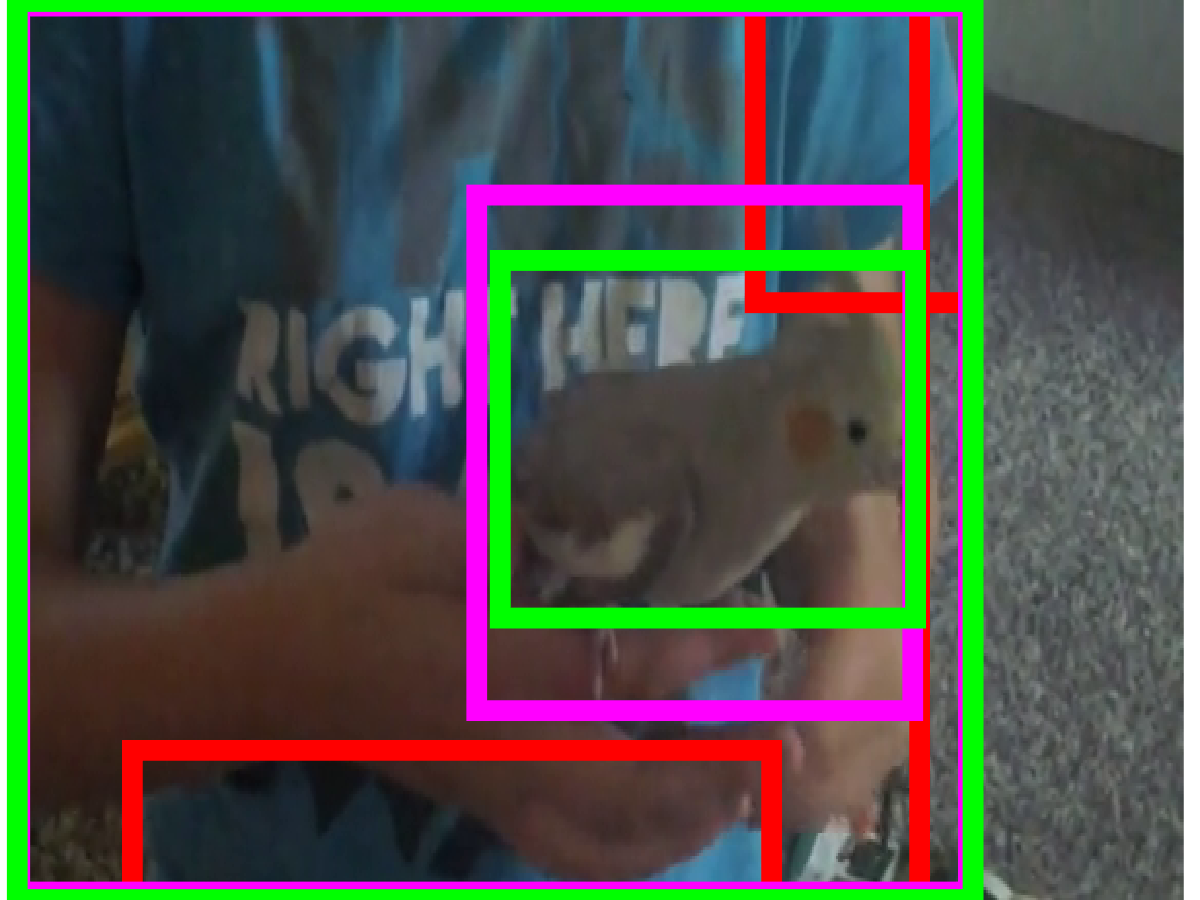} \\
\noindent \{ball, rolling\},\{dog, running\} \hspace{3cm} \{human\}, \{bird, climbing\} \\
\includegraphics[width=0.49\textwidth]{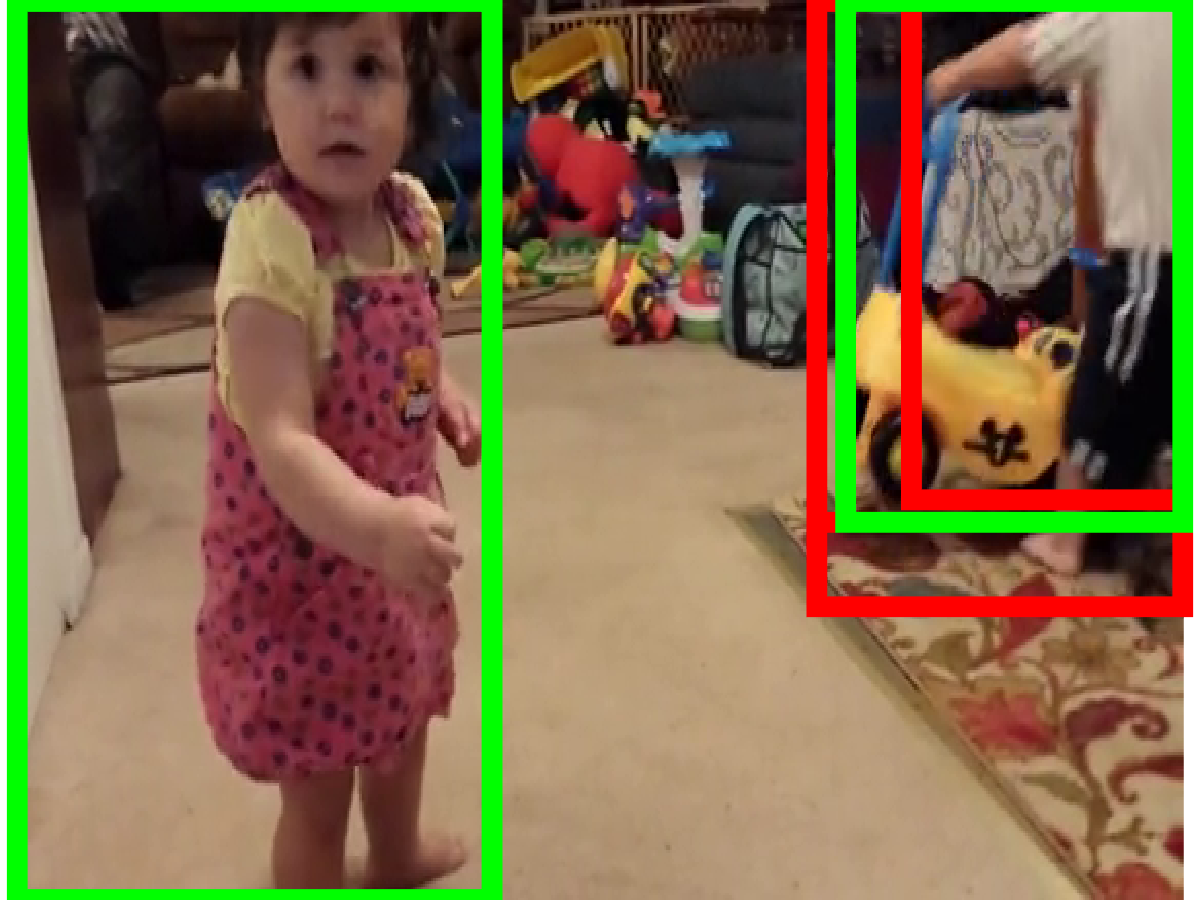} 
\includegraphics[width=0.49\textwidth]{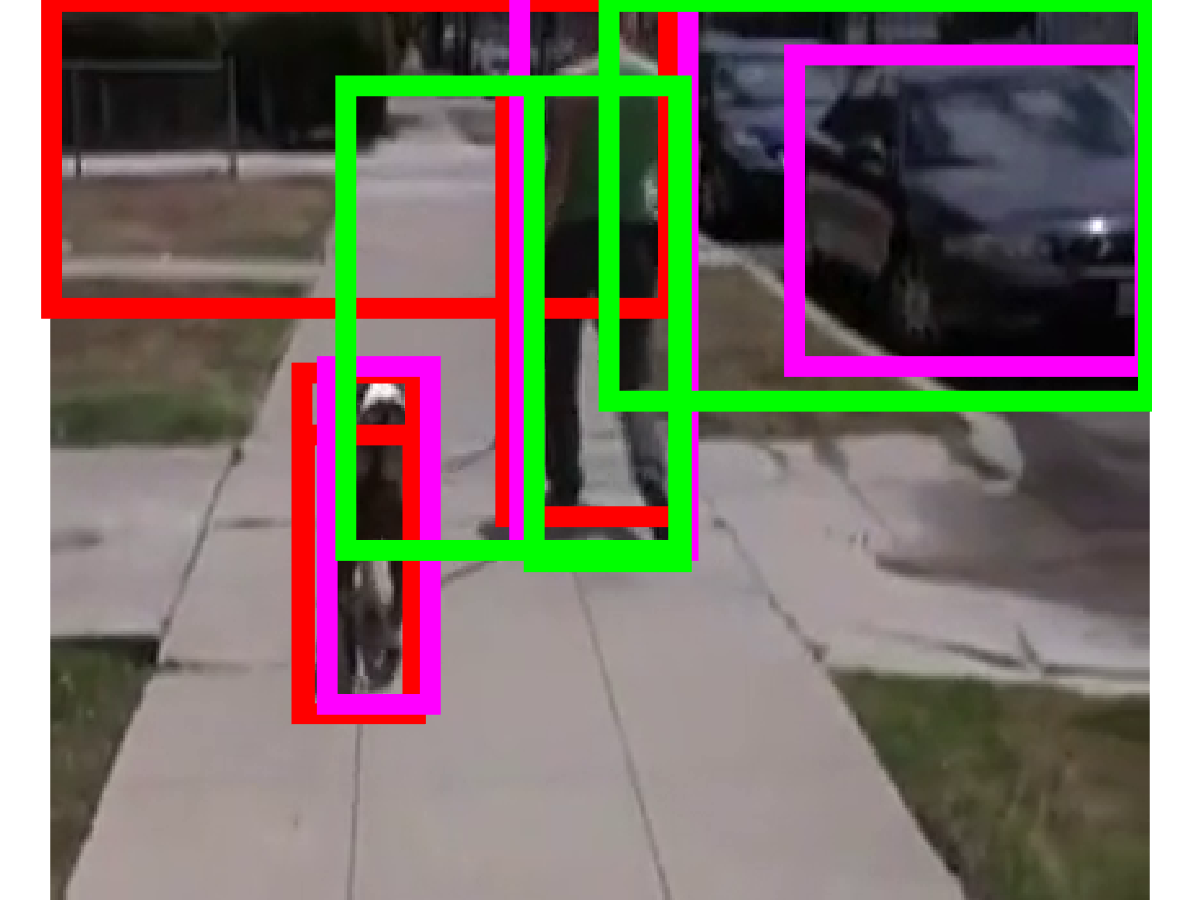} \\
\{baby, walking\}, \{human\} \hspace{2cm} \{dog, walking\}, \{human, walking\}, \{car\} \\
\includegraphics[width=0.49\textwidth]{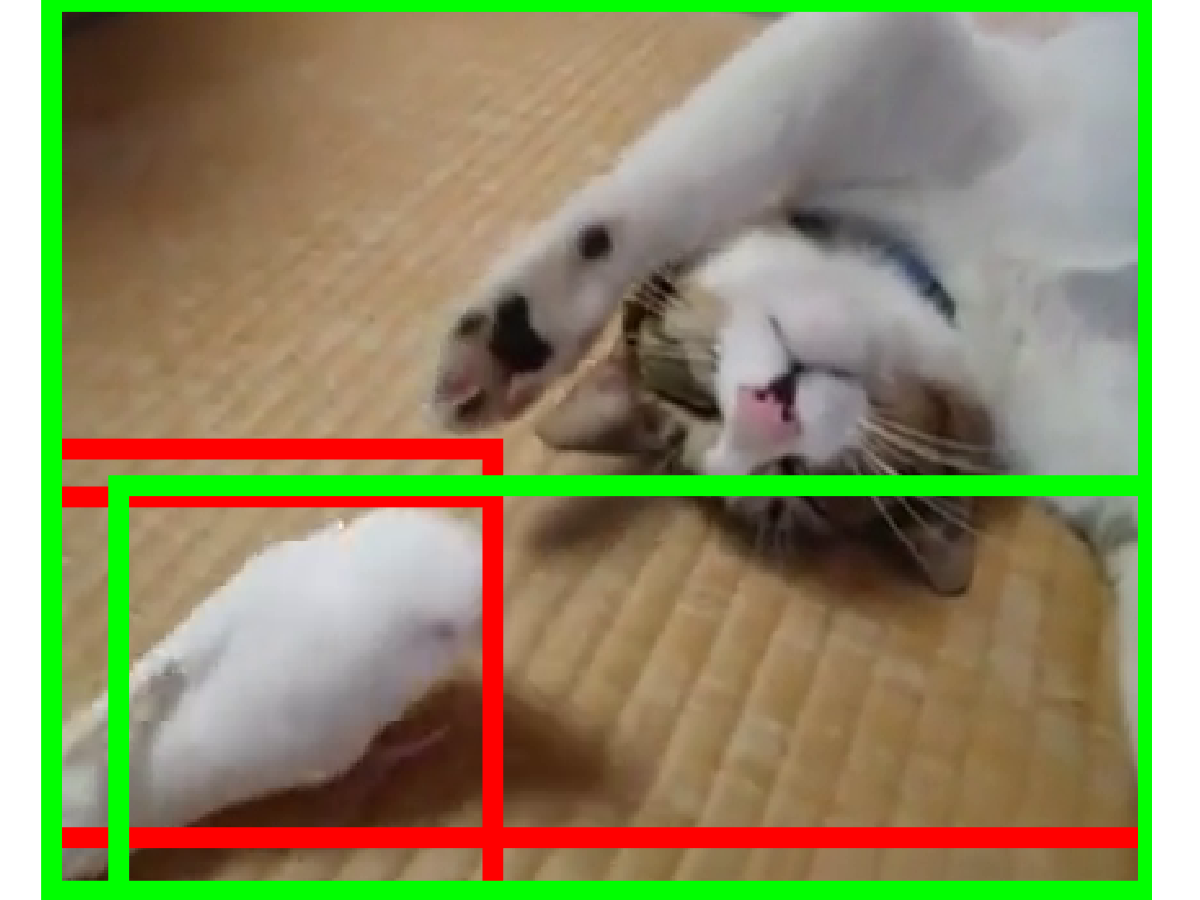} \\
\{bird, eating\},\{cat\}
\caption{Qualitative results of weakly supervised concept localization on A2D dataset using WSC-SIIBP algorithm. Tags are weak paired label input for the video.}
\label{Fig2}
\end{figure}

{\bf A2D:} Some of the additional qualitative results. In Figure \ref{Fig2}, red boxes represents generated proposals, green boxes represents the selected proposals using WSC-SIIBP algorithm and magenta boxes represents the groundtruth annotation. In case of overlapping boxes (proposals), only the last plotted rectangular box is visible. Boxes were plotted in the following order: red (first), magenta, green (last). Additionally, we have attached videos alongside this supplementary material, depicting the generated proposals and their automatic selection.

\clearpage
  
\bibliographystyle{splncs}
\bibliography{eccv2016submission}
\end{document}